\documentclass[10pt,twocolumn,letterpaper]{article}

\usepackage[pagenumbers]{cvpr} 

\definecolor{cvprblue}{rgb}{0.21,0.49,0.74}
\usepackage[pagebackref,breaklinks,colorlinks,allcolors=cvprblue]{hyperref}

\usepackage{hyperref}       
\usepackage{url}            
\usepackage{booktabs}       
\usepackage{amsfonts}       
\usepackage{nicefrac}       
\usepackage{microtype}      
\usepackage{xcolor}         
\definecolor{aliceblue}{RGB}{240,248,255}
\usepackage{colortbl}
\usepackage{amsmath} 
\usepackage{graphicx}
\usepackage{tikz,graphics,color,float,epsf,caption,subcaption}
\usepackage{booktabs, 
            makecell, multirow, tabularx}
\usepackage{algorithm}
\usepackage{algorithmic}

\newtheorem{Definition}{Definition}
\newtheorem{Theorem}{Theorem}

\def\paperID{3312} 
\def\confName{CVPR}
\def\confYear{2025}

\title{VASparse: Towards Efficient Visual Hallucination Mitigation via \\ Visual-Aware Token Sparsification}

\author{
Xianwei Zhuang\textsuperscript{1, 2}, Zhihong Zhu\textsuperscript{2}, Yuxin Xie\textsuperscript{2}, Liming Liang\textsuperscript{2}, Yuexian Zou\textsuperscript{2}\thanks{Corresponding author}\\
\small\textsuperscript{1}Guangdong Provincial Key Laboratory of Ultra High Definition Immersive Media Technology, Shenzhen Graduate School, Peking University \\
\small\textsuperscript{2}School of Electronic and Computer Engineering, Peking University \\
{\tt\small xwzhuang@stu.pku.edu.cn}\\
}

\begin{document}
\maketitle

\begin{abstract}
Large Vision-Language Models (LVLMs) may produce outputs that are unfaithful to reality, also known as visual hallucinations (VH), which significantly impedes their real-world usage.
To alleviate VH, various decoding strategies have been proposed to enhance visual information.
However, many of these methods may require secondary decoding and rollback, which significantly reduces inference speed.
In this work, we propose an efficient plug-and-play decoding algorithm via Visual-Aware Sparsification (VASparse) from the perspective of token sparsity for mitigating VH.
VASparse is inspired by empirical observations: (1) the sparse activation of attention in LVLMs, and (2) visual-agnostic tokens sparsification exacerbates VH.
Based on these insights, we propose a novel token sparsification strategy that balances efficiency and trustworthiness. 
Specifically, VASparse implements a visual-aware token selection strategy during decoding to reduce redundant tokens while preserving visual context effectively. 
Additionally, we innovatively introduce a sparse-based visual contrastive decoding method to recalibrate the distribution of hallucinated outputs without the time overhead associated with secondary decoding. 
Subsequently, VASparse recalibrates attention scores to penalize attention sinking of LVLMs towards text tokens. 
Extensive experiments across four popular benchmarks confirm the effectiveness of VASparse in mitigating VH across different LVLM families without requiring additional training or post-processing. 
Impressively, VASparse achieves state-of-the-art performance for mitigating VH while maintaining competitive decoding speed.
Code is available at \url{https://github.com/mengchuang123/VASparse-github}.

\end{abstract}    
\section{Introduction}
\label{sec:intro}

\begin{figure}[t]
   \centering
    \includegraphics[width=0.90\linewidth]{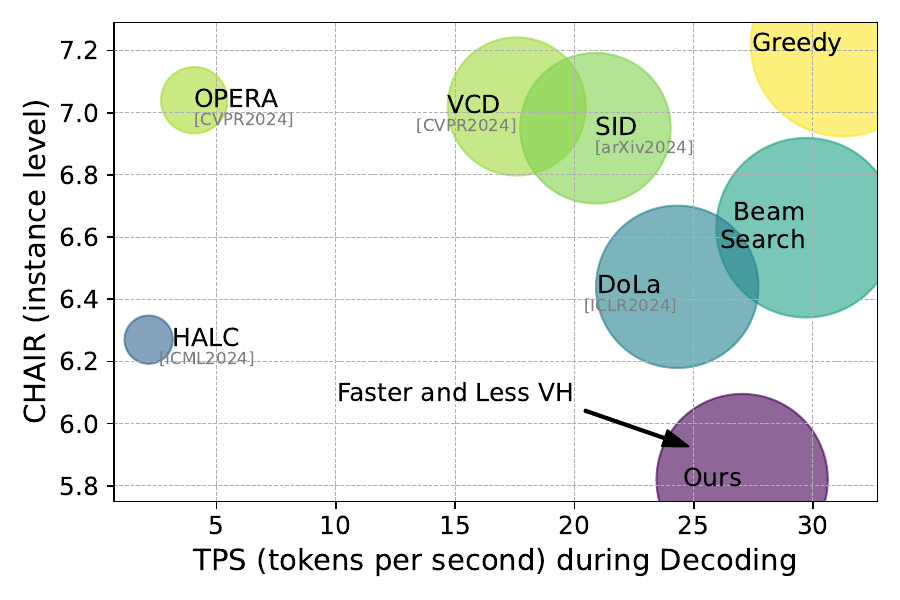}
    \caption{Comparison of decoding speed and hallucination mitigation across methods using LLaVA-1.5~\cite{Liu2023VisualIT} (max new tokens is 64), where a lower instance-level CHAIR score~\cite{Rohrbach2018ObjectHI} indicates less hallucination and higher TPS during decoding (measured by tokens generated per second) reflects greater decoding efficiency. We present the average of five runs on a single A100 GPU. Comparatively, our approach achieves both lower VH and higher efficiency.}
    \label{fig:intro}
    \vspace{-1em}
 \end{figure}

 Motivated by the success of Large Language Models (LLMs), large vision-language models (LVLMs) have made significant advancements in cross-modal understanding and generation through
 novel model architectures, training methods, and instruction-based data~\cite{Liu2023VisualIT,Gong2023MultiModalGPTAV,Li2023OtterAM,Maaz2023VideoChatGPTTD,Zhang2023VideoLLaMAAI,Zhu2023MiniGPT4EV}. 
 LVLMs excel at translating complex visual patterns into coherent language representations, leveraging the capabilities of LLMs to significantly enhance visual understanding performance and achieving impressive results across various tasks~\cite{Bai2023QwenVLAF,Dai2023InstructBLIPTG,Liu2023ImprovedBW}.
 However, LVLMs may generate outputs that inaccurately reflect the visual content provided, a phenomenon termed visual hallucinations (VH), which can affect their trustworthiness and suitability in different applications across various domains~\cite{Gunjal2023DetectingAP,Li2023EvaluatingOH,Liu2023MitigatingHI,Lovenia2023NegativeOP}.
 Additionally, recent research shows that even advanced and powerful LVLMs remain susceptible to VH~\cite{Dai2022PlausibleMN,Li2023EvaluatingOH,Guan2023HallusionBenchAA}.

Significant efforts have been directed toward mitigating VH in LVLMs to improve both the reliability and fidelity of their outputs.	
Existing strategies for reducing VH generally fall into three primary categories: post-processing and self-correction techniques~\cite{Zhou2023AnalyzingAM,Huang2023OPERAAH,Yin2023WoodpeckerHC}, instruction-based fine-tuning~\cite{Liu2023MitigatingHI,Yu2023RLHFVTT}, and decoding strategy methods~\cite{Chuang2023DoLaDB,Leng2023MitigatingOH,Chen2024HALCOH}.
Although the progressive process has been achieved, these approaches still present several significant limitations, including:
(1) a potential dependence on datasets and training, or the addition of complex post-processing steps or high-performing external LVLMs~\cite{Zhou2023AnalyzingAM,Liu2023MitigatingHI,Yu2023RLHFVTT};	
(2) the necessity for external tools and time-consuming sampling processes for visual localization~\cite{Chen2024HALCOH};
(3) multi-round decoding and repeated rollbacks significantly impact decoding speed, diminishing practical usability~\cite{Leng2023MitigatingOH, Huang2023OPERAAH}.
As illustrated in Figure~\ref{fig:intro}, such techniques may reduce VH but also compromise efficiency. For instance, state-of-the-art HALC~\cite{Chen2024HALCOH} has been shown to reduce the average decoding speed substantially.
Consequently, there is an ongoing need for more efficient solutions to mitigate VH while ensuring both efficiency and trustworthiness of LVLMs.

In this work, we present VASparse, an efficient, plug-and-play method for VH mitigation that balances efficiency and trustworthiness from the perspective of visual-aware token sparsity.
VASparse is based on several key empirical observations (\textit{cf.}  Section~\ref{sec:obs}): 
(1) the attention of LVLMs exhibits a sparse pattern;
(2) directly applying vision-agnostic sparsification methods (e.g., ~\cite{chen2024image, zhang2024sparsevlm}) for token pruning tends to worsen visual fuzziness and exacerbate VH.
Based on these insights, VASparse incorporates the following innovative strategies to balance fidelity with efficiency:

\textbf{First}, we frame the token sparsification and visual awareness in LVLMs as a unified constrained optimization problem and devise a theoretically optimal token selection strategy during decoding to solve it.
\textbf{Second}, we introduce a novel sparse-based visual contrastive decoding strategy to reduce hallucinatory tokens. Specifically, we contrast and redistribute the logits generated by visual-agnostic and visual-aware token sparsification to enhance information perception of visual entities, which utilizes embeddings to achieve logits to avoid the time overhead associated with secondary decoding. 
\textbf{Third}, we propose to penalize sinking attention using cumulative attention scores to prevent the model from overfocusing on language-biased or low-semantic tokens.

As illustrated in Figure~\ref{fig:intro}, our VASparse method achieves optimal performance in VH mitigation, with decoding speeds exceeding those of existing VH mitigation methods.
Theoretical analysis in Section~\ref{subsec:theorem} confirms the effectiveness of our visual-aware token selection strategy.
Extensive experiments across four popular VH benchmarks and three LVLM families including LLaVA-1.5~\cite{Liu2023VisualIT}, MiniGPT-4~\cite{Chen2023MiniGPTv2LL} and mPLUG-Owl2~\cite{Ye2023mPLUGOwl2RM}, demonstrate that VASparse not only delivers superior performance but also achieves competitive decoding speeds (e.g., achieving better performance and up to 12.9 $\times$ speed improvement than HALC~\cite{Chen2024HALCOH}).

In summary, our main contributions are threefold:
\begin{itemize}
\item We explore VH mitigation from the perspective of token sparsification during decoding and present a novel, efficient, plug-and-play approach that achieves both model fidelity and efficiency, which unifies token sparsity and visual-aware enhancement as an optimization problem.
\item We propose a novel visual-aware token selection strategy, along with a sparse-based visual contrastive decoding method to alleviate VH which utilizes embeddings to achieve contrasted logits and avoids multi-round decoding.
\item Comprehensive experiments and evaluations demonstrate that VASparse significantly outperforms existing VH mitigation methods in both performance and decoding speed.
\end{itemize}

\begin{figure*}
   \centering
   \begin{subfigure}{0.33\linewidth}
     \includegraphics[width=1\linewidth]{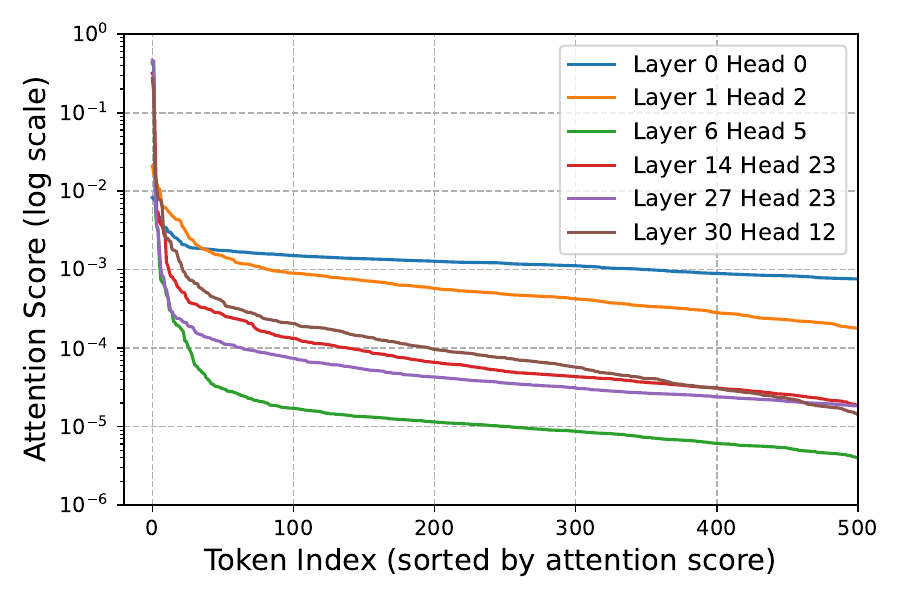}
     \caption{Attention between Tokens is Highly Sparse.}
     \label{fig:short-a}
   \end{subfigure}
   \hfill
   \begin{subfigure}{0.33\linewidth}
     \includegraphics[width=1\linewidth]{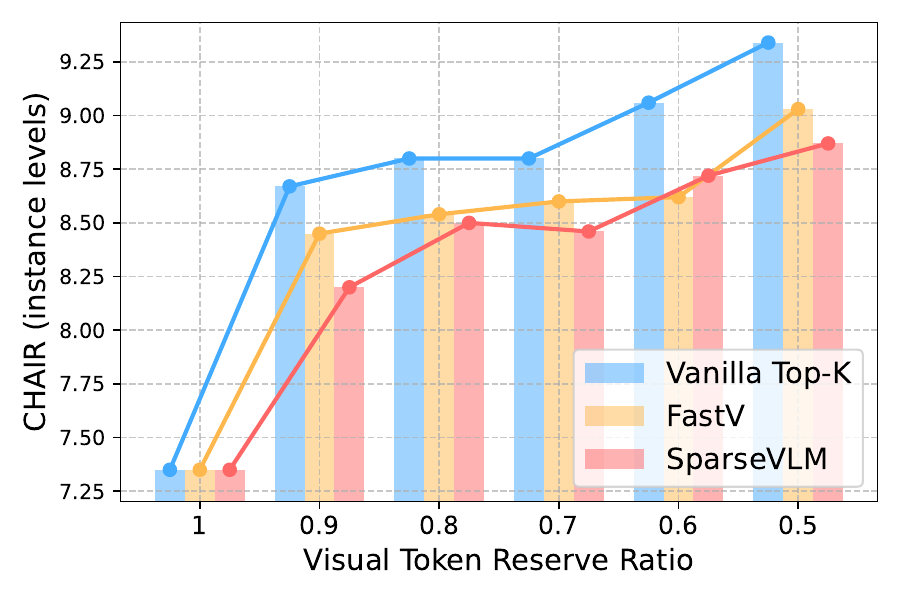}
     \caption{Visual-Agnostic Token Sparsification and VH.}
     \label{fig:short-b}
   \end{subfigure}
     \hfill
   \begin{subfigure}{0.33\linewidth}
     \includegraphics[width=1\linewidth]{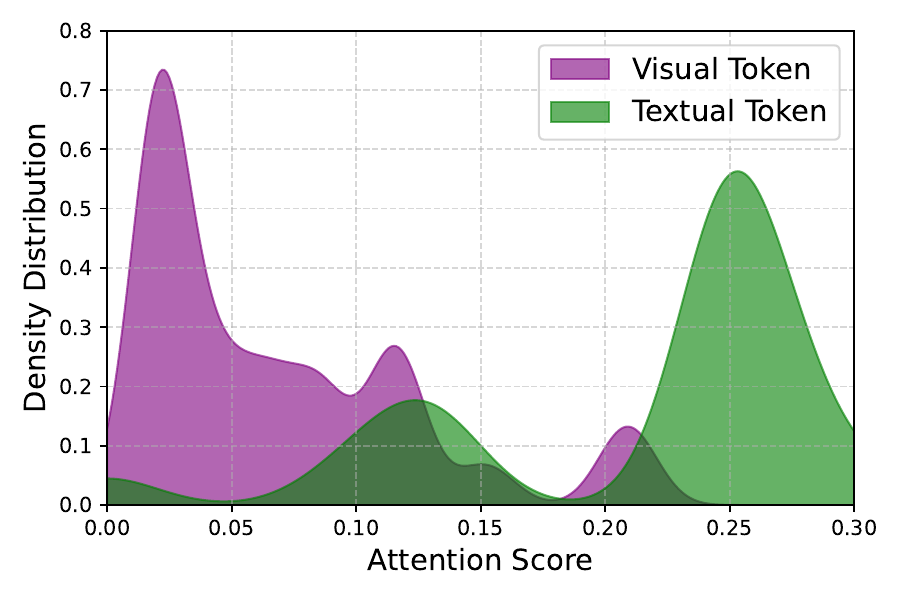}
     \caption{Attention Density of Visual and Textual Tokens.}
     \label{fig:short-c}
   \end{subfigure}
   \caption{VH evaluation and attention analysis using LLaVA-1.5 on the CHAIR benchmark: (a) token sorting by attention score; (b) token sparsification effects observed with Vanilla Top-K, FastV~\cite{chen2024image}, and SparseVLM~\cite{zhang2024sparsevlm} on sampled 500 images from the MSCOCO validation set, where Vanilla Top-K denotes keeping tokens with top-K scores in $1$-th layer; and (c) attention density distribution across various tokens.}
   \vspace{-1em}
   \label{fig:short}
 \end{figure*}

\section{Related Work}
\label{sec:related}

\textbf{Large Vision-Language Model.}~
In recent years, significant progress has been made in visual understanding~\cite{zhao2024graco,zhao2024detrsbeatyolosrealtime} and question answering~\cite{fkdsKDProD,xie2024gpa,zhuang2024towards,yin2025atri}.
Recent efforts have attempted to employ NLP methods and LLMs~\cite{touvron2023llama, touvron2023llama-2, chiang2023vicuna,ru2025reallyfilterrandomnoise, Zhuang_Cheng_Zou_2024,taori2023stanford,zhuang2025unicott,zhuang2025vargptunifiedunderstandinggeneration} as text decoders, combined with visual decoders~\cite{radford2021learningtransferablevisualmodels} and a projector, to construct high-performing LVLMs.
By integrating visual information with user instructions, LVLMs have achieved significant performance in generating diverse responses and handling complex visual understanding tasks.
LLaVA~\cite{liu2024visual} and LLaVA-1.5~\cite{liu2024improved} integrate pretrained visual encoders and text decoders, leveraging instruction fine-tuning to achieve strong multimodel understanding performance.
InstructBLIP~\cite{dai2023instructblipgeneralpurposevisionlanguagemodels} and MiniGPT-4~\cite{zhu2023minigpt} utilize a Q-former~\cite{li2023blip} to aggregate multimodal features, thereby reducing the number of visual tokens required.
With optimized architectures, training modes, and diverse data, increasingly advanced LVLM families, such as Qwen-VL~\cite{bai2023qwen}, mPLUG-Owl2~\cite{ye2024mplug}, and InternVL~\cite{chen2024internvl}, have also been proposed.
In this work, we use various architectures of LLaVA-1.5~\cite{liu2024improved}, MiniGPT-4~\cite{zhu2023minigpt}, and mPLUG-Owl2~\cite{ye2024mplug} to evaluate our approach for mitigating VH.

\textbf{VH and Evaluation.}~
LVLMs face challenges from VH which specifically refers to instances where generated content includes inaccurate object descriptions or is unfaithful to the input image information. 
This phenomenon has been observed in both early BERT-based models~\cite{Li2019VisualBERTAS} and recent LVLMs~\cite{Maaz2023VideoChatGPTTD, Zhang2023VideoLLaMAAI, Zhu2023MiniGPT4EV}. 
In the realm of LVLMs, extensive research has delved into the evaluation and detection of VH~\cite{Li2023EvaluatingOH, Wang2023EvaluationAA, Lovenia2023NegativeOP}. 
CHAIR~\cite{Rohrbach2018ObjectHI} is one of the most widely adopted benchmarks for assessing VH. 
POPE~\cite{Li2023EvaluatingOH} evaluates VH through a binary classification framework, utilizing precision, recall, and accuracy. 
Furthermore, HALC~\cite{Chen2024HALCOH} proposes an offline POPE (OPOPE) to enhance VH evaluation. 
And MME~\cite{Fu2023MMEAC} provides a comprehensive performance assessment of LVLMs with respect to objects, attributes, and other factors. 
We combine these metrics with decoding speed to comprehensively evaluate the effectiveness of our VASparse in reducing VH while maintaining high efficiency.

\textbf{VH Mitigation.}~
To mitigate VH, various strategies have been developed. 
Current efforts for reducing VH generally fall into three categories: post-processing techniques~\cite{Zhou2023AnalyzingAM,Huang2023OPERAAH} and self-correction methods~\cite{Yin2023WoodpeckerHC}; human feedback-based methods~\cite{Liu2023MitigatingHI,Yu2023RLHFVTT}; and decoding strategy approaches~\cite{Chuang2023DoLaDB,Leng2023MitigatingOH,Chen2024HALCOH,zhuang-etal-2024-game}. 
However, the first two strategies may require additional datasets and training or the integration of more powerful external LVLMs~\cite{Zhou2023AnalyzingAM,Liu2023MitigatingHI,Yu2023RLHFVTT}. 
The third approach~\cite{Chuang2023DoLaDB,Leng2023MitigatingOH,Chen2024HALCOH,Leng2023MitigatingOH,Huang2023OPERAAH,huo2024selfintrospectivedecodingalleviatinghallucinations} primarily explores contrastive decoding strategies based on visual comparisons, which may involve multiple rounds of decoding, time-consuming rollbacks, or even the use of external detection tools. 
Our work focuses on designing efficient, plug-and-play methods that require no additional training.
 
\section{Observation and Motivation}
\label{sec:obs}

\begin{figure}[t]
  \centering
   \includegraphics[width=0.7\linewidth]{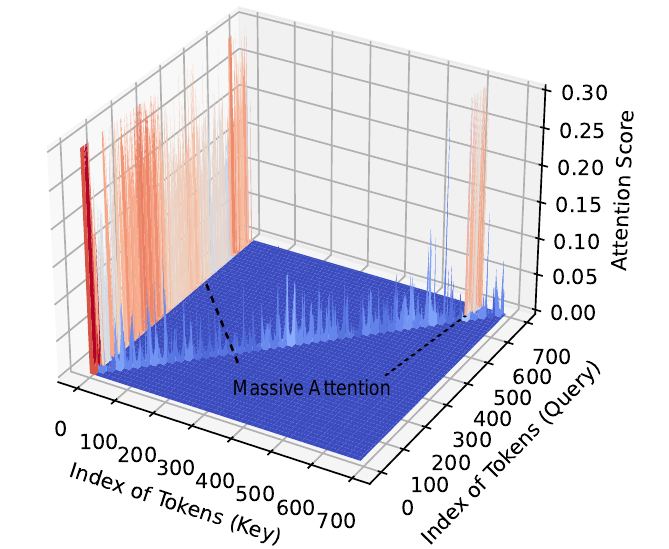}
   \caption{Attention sinking phenomenon in LVLMs: in the 8-th layer and 26-th attention head of LLaVA-1.5, exhibits a substantial concentration of attention on specific tokens, e.g., \texttt{<.>} and \texttt{<s>}.}
   \label{fig:onecol}
   \vspace{-1em}
\end{figure}

\begin{figure*}[t]
  \centering
   \includegraphics[width=1\linewidth]{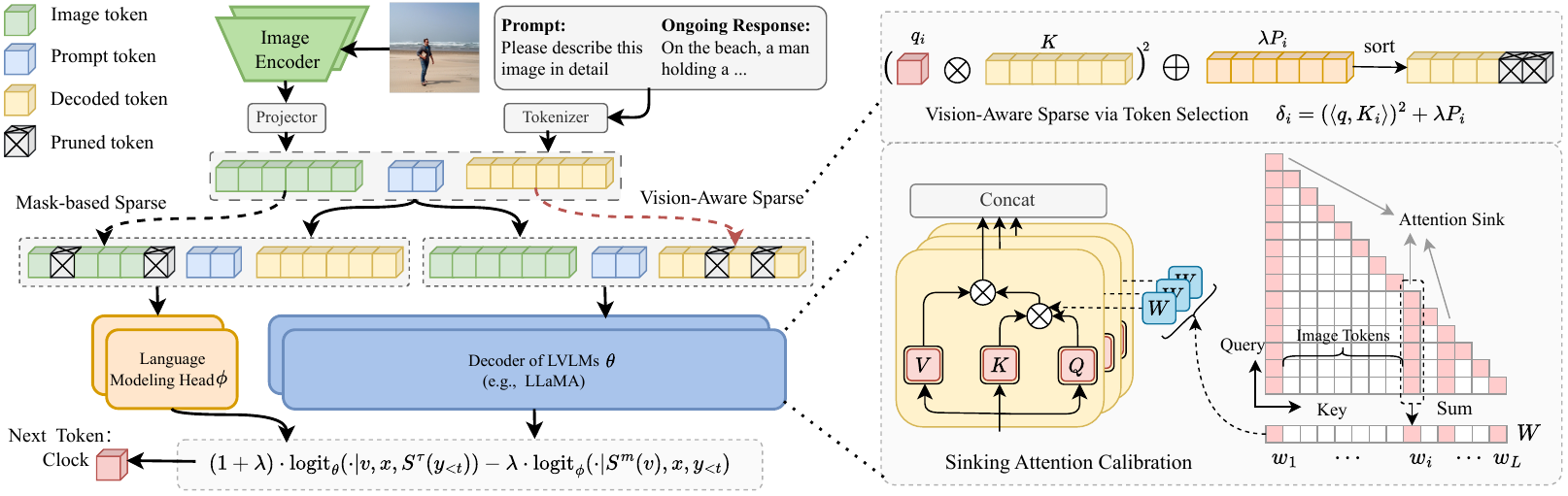}

   \caption{The illustration of the proposed VASparse framework, which consists of (1) the visual-aware token selection designed 
   to prune the generated tokens during decoding; (2) a sparse-based visual contrastive decoding method to recalibrate the distribution of hallucinated outputs;
   and (3) the calibration strategy for punishing sinking attention.}
   \label{fig:main}
\end{figure*}

In this section, we present the motivation behind our VASparse for mitigating VH. 
We first provide evidence of attention sparsity in LVLMs and observe that vision-agnostic sparsification methods can intensify VH. 
Additionally, we emphasize the necessity of attending to image tokens and applying penalties to tokens prone to attention sinking.

\subsection{Sparse Activation in LVLM Attention}
\label{subsec:obs-1}
\textbf{Observation}:~
As shown in Figure~\ref{fig:short-a}, we sort the attention scores calculated for decoding tokens of LVLMs in ascending order. 
We observe that the attention scores exhibit a clear long-tail distribution, with only a small portion of tokens being heavily activated during decoding. 
Our results in Figure~\ref{fig:short-a} indicate that retaining only the top 1\% of tokens with the highest attention scores can recall over 98\% of the total attention score.
This suggests that attention in most layers of the LVLM decoder is sparse. 

\textbf{Insights}:~
Our findings substantiate that self-attention in most layers of the LVLM decoder is sparse. This insight suggests the potential for pruning corresponding tokens to reduce computational cost during decoding.

\subsection{Vision-Agnostic Sparsification Aggravates VH}
\label{subsec:obs-2}
\textbf{Observation}:~
Given the sparsity of attention in LVLMs, we evaluate VH with vision-agnostic (do not adjust token selection during decoding) token sparsification, including the vanilla Top-K strategy, FastV~\cite{chen2024image} and SparseVLM~\cite{zhang2024sparsevlm}. 
As shown in Figure~\ref{fig:short-b}, we observe that as the level of sparsification increases, the model becomes more prone to VH.

\textbf{Insights}:~
Our empirical findings indicate that these vision-agnostic sparsification techniques exacerbate VH in LVLMs, suggesting that merely applying such methods to speed up decoding may undermine output trustworthiness.

\subsection{Distinct Distribution of Image and Text Tokens} 
\label{subsec:obs-3}
\textbf{Observation}:~
We analyze the attention distribution of visual and textual tokens, with the results shown in Figure~\ref{fig:short-c}. A clear divergence in distribution is evident: image tokens predominantly occupy lower-attention regions, whereas text tokens concentrate in higher-attention regions.

\textbf{Insights}:~
These findings suggest that LVLMs tend to prioritize text tokens over image tokens during decoding. 
This explains why vision-agnostic token sparsification strategies may worsen hallucinations (\textit{cf.} Section~\ref{subsec:obs-2}): they are more likely to prune low-attention image tokens, which may contain crucial visual information. 
This insight highlights the potential benefits of improving the model's awareness of image tokens during sparsification.

\subsection{Attention Sinking on Textual Tokens}
\label{subsec:obs-4}
\textbf{Observation}:~
We further analyzed the attention patterns in LVLMs and observed a significant attention "sink" effect~\cite{xiao2024efficientstreaminglanguagemodels,Huang2023OPERAAH} in certain text tokens (as illustrated in Figure~\ref{fig:onecol}). 
This phenomenon resembles the summary token and attention bias effects observed in LLMs~\cite{xiao2024efficientstreaminglanguagemodels}. 
However, distinct from LLMs, our findings indicate that in LVLMs, attention sink tokens are primarily concentrated in text tokens, even when text tokens are vastly outnumbered by image tokens. 
Notably, these attention sink tokens are typically low in semantic content, such as \texttt{<.>} and \texttt{<s>}.

\textbf{Insights}:~
Tokens with attention sinking in LVLMs exhibit high attention and low semantic information. This pattern suggests an intrinsic bias within LVLMs. However, excessive focus on low-semantic tokens may cause the model to rely heavily on linguistic priors and neglect visual information. Therefore, applying penalties to these sinking tokens could enhance the LVLM's perception of visual tokens.

\section{Methodology}
\label{sec:methods}

\subsection{Preliminaries}
We consider a general LVLM $\theta$, which integrates a vision encoder, a vision-text interface, and a decoder of LLM.
Initially, the image $v$ undergoes processing through the vision encoder to produce embeddings, which are then modified by the interface (e.g., linear layer and Q-Former~\cite{li2023blip}) to align with the query $x$. The combined data serves as input to the decoder, which autoregressively generates the output $y$ as:
\begin{equation}
\begin{aligned}
    y_t &\sim p_{\theta}(y_t | v, x, y_{<t})\propto \exp\left( \operatorname{logit}_{\theta}(y_t | v, x, y_{<t}) \right),
\end{aligned}
\end{equation}
where $y_t$ represents the $t$-th token of $y$, and $y_{<t}$ refers to the sequence of tokens generated prior to the $t$-th step. The function $\operatorname{logit}_{\theta}$ is the logit distribution function.

During decoding, the key $K$ and value $V$ within the attention head are derived from preceding decoding steps and stored in a key-value cache to avoid redundant computations. Consequently, the attention with dimension $D$ for decoding the $t$-th token proceeds during decoding as follows:
\begin{equation}
\operatorname{Attention}(q_t, K_{\leq t}) = \operatorname{Softmax}\left(\frac{q_t K_{\leq t}^\top}{\sqrt{D}}\right),
\end{equation}
where $q_t$ is the query for the current decoding step, and $K_{\leq t}$ represents the keys up to and including step $t$.

Our primary goal is to reduce generated hallucinatory tokens to preserve the trustworthiness of the generated text and maintain efficient decoding speed.

\subsection{Problem Formulation}
\label{subsec:Problem}
Building on our observations in Section~\ref{sec:obs}, we decompose the unified objective of achieving both trustworthiness and efficiency for LVLMs into the following sub-goals:

\textbf{Goal 1 (Token Sparsification)}:~
Given the sparsity of LVLMs (\textit{cf.} Section~\ref{subsec:obs-1}),
we define token sparsification through a binary mask $M$, where each element $M_i \in \{0, 1\}$. Optimal sparsification minimizes $\sum_{i=1}^L M_i$ while maximizing the recall of attention scores, aiming for $q({M} \odot K)^\top$ to approximate the full attention score $qK^\top$ as closely as possible, where $L$ is the generated sequence length and $M_i = 0$ indicates that the token $K_i$ will be pruned during decoding.

\textbf{Goal 2 (Vision-Aware Decoding)}:~
During decoding, some tokens may hold lower attention scores but are crucial for decoding visually relevant instances. Ignoring these tokens can exacerbate VH (\textit{cf.} Section~\ref{subsec:obs-2} and \ref{subsec:obs-3}). We assign each token a vision-aware saliency score $P_i$ to represent its importance for decoding visual instances. A higher $P_i$ indicates that the token should be more likely to be retained.

The above objectives can be summarized as maintaining the original attention scores as much as possible while sparsifying the tokens and considering visual information during the decoding process. 
We innovatively unify these optimization goals into a constrained optimization problem which minimizes the error between the recalled attention scores and the full attention scores: 

\begin{Definition}
    \textbf{(Unified Objective)}:
    We define the joint objective of trustworthiness and efficiency in LVLMs as the solution to the following constrained optimization problem:
    \begin{equation}
    \begin{aligned}
    \min_{M} \quad \mathcal{E}(M) &= {\left\| qK^\top - q ({M} \odot K)^\top \right\|^2} - {\lambda P \cdot M} \\
    = \sum_{i=1}^L &  \left( \langle q, K_i \rangle - {M_i} \langle q, K_i \rangle \right)^2 - \lambda P_i \cdot M_i \\
    \text{s.t.} \quad M_i \in &\{0, 1\},  \forall i = 1, 2, \dots, L; \quad
    \sum_{i=1}^L M_i = S,
    \end{aligned}
    \end{equation}
    where, $q \in \mathbb{R}^{1 \times D }$, $K_i \in K$ and $K_i \in \mathbb{R}^{1 \times D }$, $||\cdot||^2$ represents the $L_2$ norm. $\langle \cdot,\cdot \rangle$ denotes the inner product, and $S$ is the sparsity rate, and $\lambda$ is a tradeoff parameter used to balance visual perception and attention recall.

\label{def:objective}
\end{Definition}

The objective~\ref{def:objective} inherently includes the following constraints:
(1) Sparsity Constraint: $\sum_{i=1}^{L} M_i  = S$, and $S$ denotes the number of non-zero elements in $M$, with $S < L$ and $M_i \in \{0, 1\}$; (2) Visual Saliency Constraint: $P = \{P_i\}_{i=1}^L$ represents the visual-aware scores.
To solve this problem~\ref{def:objective} efficiently, we propose a novel \textbf{visual-aware token selection strategy} to achieve efficient VH mitigation as the overall framework shown in Figure~\ref{fig:main}.

\subsection{Visual-Aware Token Selection}
\label{subsec:selection}
To solve the unified objective (\textit{Def}. \ref{def:objective}) and mitigate VH efficiently, we propose a visual-aware token selection strategy. Specifically, for each attention head, we rank tokens based on an aggregated score $\delta_i$ in descending order, and setting $M_i = 1$ for the top-$S$ tokens and $M_i = 0$ for the rest. The proposed aggregation score $\delta_i$ for each token is defined as:
\begin{equation}
    \delta_i = \left( \langle q, K_i \rangle \right)^2 + \lambda P_i,
    \label{eq:select}
\end{equation}
where, $\langle \cdot,\cdot \rangle$ denotes the inner product, the score $\delta_i$ combines both the attention score $\langle q, K\rangle$ and the visual saliency $P_i$, ensuring that the visually relevant tokens are retained while preserving computational efficiency.

To obtain visual-aware scores (Goal 2 in Section~\ref{subsec:Problem}), we utilize the attention scores of each generated token and the image tokens, which are treated as the visual saliency scores for the respective tokens. 
Specifically, we compute the visual saliency score $P$ by retaining the weights from the last attention head in the LVLM's historical calculations:
\begin{equation}
P_i = \frac{\exp \left({\sum_{k \in \mathcal{I}(v)} a_{i,k}}\right)}{\sum_{j} \exp\left({\sum_{k \in \mathcal{I}(v)} a_{j,k}}\right)},
\end{equation}
where $\mathcal{I}(v)$ represents the set of image tokens and $a_{i,j}$ is the attention score between tokens $i$ and $j$.

By using the image token attention scores as a measure of significance, we can effectively leverage the attention weights already computed, while avoiding the introduction of additional computational overhead.
For the discarded token set $\mathcal{T} = \{ K_i \mid M_i = 0 \} $, we employ the $k$-nearest neighbor density peak aggregation algorithm~\cite{rodriguez2014clustering} to achieve adaptive token aggregation.
Tokens within the same cluster are summed and retained as a single aggregated token.

\subsection{Sparse-based Visual Contrastive Decoding} 
Based on our empirical observations, we can leverage the finding that vision-agnostic token sparsification intensifies VH to mitigate language bias in the output distribution.
We innovatively propose to amplify the informational contrast within the visual context by redistributing logits in the output by contrasting the decoding probability distributions of vision-aware and vision-agnostic (mask-based) sparsifications $S^{\tau}$ and $S^{m}$.
However, directly using the output distribution from LVLMs to obtain the contrastive logit distribution would inevitably incur significant overhead due to the secondary decoding process.
To address this, we propose using only the embeddings of vision-agnostic tokens as input to the language decoding head $\phi$ of the LLM decoder to obtain the logit distribution, without going through the full text decoder.
Specifically, we adopt the proposed visual-aware sparsification strategy (\textit{cf.} Section~\ref{subsec:selection}) to obtain the logit distribution $\operatorname{logit}_\theta$. 
Then, we randomly mask the visual tokens and input their embeddings directly into the language decoding head of the LLM to obtain the contrastive logit distribution $\operatorname{logit}_\phi$.
Finally, we assign the logit distributions of the tokens to obtain the final results:
\begin{equation}
    \begin{split}
    y_t \sim (1+\alpha) &\cdot \operatorname{logit}_\theta \left(\cdot \mid v, x, S^{\tau}(y_{<t})\right) \\ 
    -\alpha& \cdot\operatorname{logit}_\phi\left(\cdot \mid  S^{m}(v), x, y_{<t}\right),
    \end{split}
    \label{eq:sparse-con}
\end{equation}
where, $\alpha$ is a trade-off.
Note that our decoding strategy bypasses the LVLM's decoder (e.g., a LLaMA2-7B~\cite{touvron2023llama-2}), thereby avoiding the secondary computational overhead.
Inspired by~\cite{Leng2023MitigatingOH}, we apply adaptive plausibility constraints to our sparse-based visual contrastive decoding.

\begin{table*}[htbp]
    \centering
        \renewcommand{\arraystretch}{1.0}
    \tabcolsep=0.25cm
    \small
      \begin{tabular}{l|ccc|ccc|ccc}
      \toprule
       \multirow{2}[2]{*}{\textbf{Methods}} & \multicolumn{3}{c|}{\textbf{LLaVA-1.5}} & \multicolumn{3}{c|}{\textbf{MiniGPT-4}} & \multicolumn{3}{c}{\textbf{mPLUG-Owl2}} \\
            & CHAIR$_i\downarrow$ & CHAIR$_s\downarrow$ & TPS$\uparrow$   & CHAIR$_i\downarrow$ & CHAIR$_s\downarrow$ & TPS$\uparrow$   & CHAIR$_i\downarrow$ & CHAIR$_s\downarrow$ & TPS$\uparrow$ \\
      \midrule
       FastV$^*$ & 8.53  & 26.76  & 33.21  & 16.72  & 41.32  & 38.29  & 11.40  & 38.49  & 24.6 \\
            SparseVLM$^*$  & 8.44  & 26.11  & 32.47  & 16.38  & 40.93  & 37.81  & 11.35  & 38.99  & 23.73 \\
            Woodpecker$^\dagger$  & 6.72  & 19.79  & -     & 12.09  & 31.69  & -     & 8.99  & 25.05  & - \\
            LURE$^\dagger$  & 6.67  & 19.75  & -     & 11.80  & 31.67  & -     & 7.78  & 22.53  & - \\
            Greedy & 7.22  & 22.20  & 31.25  & 12.17  & 31.47  & 36.64  & 8.94  & 24.42  & 20.36 \\
            Beam Search & 6.43  & 19.97  & 29.91  & 11.57  & 31.80  & 32.27  & 8.72  & 23.87  & 19.62 \\
            OPERA & 7.04  & 21.28  & 4.36  & 12.34  & 32.63  & 5.57  & 9.07  & 24.48 & 3.56 \\
           VCD   & 7.02  & 21.40  & 17.58  & 11.90  & 30.60  & 17.69  & 9.13  & 24.89  & 9.89 \\
            DoLa  & 6.44  & 20.23  & 23.61  & 11.62  & 30.58  & 25.01  & 8.88  & 24.67  & 14.74 \\
            SID   & 6.95  & 20.83  & 20.88  & 11.85  & 31.73  & 22.95  & 8.54  & 23.55  & 12.95 \\
            HALC  & 6.27  & 19.64  & 2.15  & 11.69  & 31.76  & 3.86  & 7.71  & 23.48  & 1.52 \\
            \midrule
            \rowcolor{aliceblue} Ours  & \textbf{5.82 } & \textbf{18.51 } & 27.73  & \textbf{11.35}  & \textbf{30.19}  & 30.87  & \textbf{7.36}  & \textbf{22.03}  & 18.18 \\
      \bottomrule
      \end{tabular}%
    \caption{Comparison of the average CHAIR evaluation results (instance levels CHAIR$_i$ and sentence levels CHAIR$_s$ )and token per second (TPS) during decoding with different baselines on MSCOCO datasets of five random runs, with whole statistical results in Appendix. $^*$ represents the image token sparsity method and $\dagger$ is the post-hoc methods.}
    \label{tab:main-1}%
    \vspace{-0.5em}
\end{table*}%

\subsection{Sinking Attention Penalty} 
\label{subsec:attn}
Our observations (\textit{cf.} Section~\ref{subsec:obs-4}) indicate a pronounced attention sinking in LVLMs, where tokens receive disproportionately high attention scores despite low semantic information. 
Excessive focus on such tokens can blur visual information during decoding. Therefore, a targeted penalty should be applied to tokens exhibiting abnormally high attention scores. We define a penalty weight matrix $W = \{w_1, \cdots, w_L\}$, where each $w_i$ serves as a penalty factor for anomalous attention scores.
To efficiently implement the penalty for sinking attention, we accumulate the attention scores of each token with subsequent queries to evaluate the degree of sinking. 
We then apply $softmax$ normalization to obtain a calibration weight for sinking attention:
\begin{equation}
w_j = \frac{\exp\left( \sum_{i=j}^{L} a_{i,j} \right)}{\sum_{k=1}^{L} \exp\left( \sum_{i=k}^{L} a_{i,k} \right)},
\label{eq:penilty}
\end{equation}
where $a_{i,j} $ denotes the element in the $i$-th row and $j$-th column of the attention matrix, and $w_j$ represents the $j$-th element of the weight vector $W$ after applying the $softmax$ operation. 
This approach ensures that sinking attention is evaluated progressively across subsequent queries, and $W$ will be utilized as a weight as $(1+\beta)q K ^ {\top} - \beta W \odot q K ^ {\top}$ during decoding, as shown in Figure~\ref{fig:main}.

\subsection{Theoretical Analysis} 
\label{subsec:theorem}
\begin{Theorem}
\textbf{(Global Optimality)}: By employing the selection strategy defined in Section~\ref{subsec:selection}, we can obtain a globally optimal solution for the optimization problem defined in \textit{Def}.~\ref{def:objective}. Specifically, the sparse mask $M$ derived from this selection strategy satisfies:
\begin{equation}
    M^* = \arg \min_{M} \mathcal {E}(M).
\end{equation}
\label{the:global}
\end{Theorem}
\vspace{-5mm}
\textbf{Intuition:}~
The proof and more analysis of the theorem~\ref{the:global} is provided in the Appendix.
This theorem ensures that the proposed token selection strategy yields the minimum error $\mathcal{E}(M)$. 
This theoretical analysis further validates the effectiveness of the proposed VASparse in achieving both token sparsification and efficient visual perception.
\section{Experiments}
\label{sec:experiments}

\begin{table*}[htbp]
    \centering
            \renewcommand{\arraystretch}{1.05}
    \tabcolsep=0.25cm
    \small
      \begin{tabular}{l|ccc|ccc|ccc}
      \toprule
      \multirow{2}[2]{*}{\textbf{Methods}} & \multicolumn{3}{c|}{\textbf{LLaVA-1.5}} & \multicolumn{3}{c|}{\textbf{MiniGPT-4}} & \multicolumn{3}{c}{\textbf{mPLUG-Owl2}} \\
            & \textit{Random} & \textit{Popular} & \textit{Adversarial} & \textit{Random} & \textit{Popular} & \textit{Adversarial} & \textit{Random} & \textit{Popular} & \textit{Adversarial} \\
      \midrule
      Woodpecker$^\dagger$ & 59.73  & 58.53  & 58.07  & 53.84  & 51.70  & 51.27  & 58.10  & 53.07  & 55.42  \\
      LURE$^\dagger$  & 60.08  & 58.63  & 58.34  & 53.91  & 52.37  & 51.38  & 58.28  & 53.15  & 55.65  \\
      Greedy & 58.75  & 57.42  & 56.64  & 53.71  & 51.68  & 51.92  & 57.40  & 53.43  & 55.43  \\
      Beam Search & 60.38  & 58.98  & 58.43  & 53.97  & 52.27  & 51.93  & 55.31  & 52.89  & 53.12  \\
      OPERA & 59.80  & 58.42  & 58.00  & 53.08  & 51.32  & 51.20  & 55.70  & 53.41  & 53.66  \\
      VCD   & 60.05  & 58.34  & 58.02  & 53.26  & 51.50  & 51.07  & \textbf{58.63 } & 54.87  & 56.13  \\
      DoLa  & 59.36  & 58.08  & 57.44  & 53.83  & 51.93  & 51.72  & 57.21  & 53.38  & 55.24  \\
      SID   & 61.63  & 59.62  & 58.83  & 53.86  & 51.98  & 51.77  & 55.82  & 53.46  & 56.07  \\
      HALC  & 60.46  & 59.33  & 58.50  & 53.93  & 52.06  & 51.80  & 56.29  & 53.38  & 55.84  \\
      \midrule
      \rowcolor{aliceblue} Ours  & \textbf{62.13 } & \textbf{60.93 } & \textbf{59.20 } & \textbf{54.87 } & \textbf{52.93 } & \textbf{52.70 } & 58.27  & \textbf{55.28 } & \textbf{56.77 } \\
      \bottomrule
      \end{tabular}%
    \caption{Comparison of the average F1-score evaluation results under different settings (i.e., \textit{ Random, Popular, Adversarial}) with different baselines and our VASparse on offline POPE benchmark~\cite{Li2023EvaluatingOH, Chen2024HALCOH} of five random runs, with whole statistical results in Appendix. Higher F1-score indicate better performance and bold indicates the best results. $\dagger$ denotes the post-hoc method.}
    \label{tab:main-2}%

  \end{table*}%

\begin{table*}[htbp]
    \centering
        \renewcommand{\arraystretch}{1.05}
    \tabcolsep=0.15cm
    \small    
      \begin{tabular}{l|cccc|cccc|cccc}
      \toprule
      \multirow{3}[4]{*}{\textbf{Methods}} & \multicolumn{4}{c|}{\textbf{LLaVA-1.5}} & \multicolumn{4}{c|}{\textbf{MiniGPT-4}} & \multicolumn{4}{c}{\textbf{mPLUG-Owl2}} \\
  \cmidrule{2-13}          & \multicolumn{2}{c}{\textbf{Object-level}$\uparrow$ } & \multicolumn{2}{c|}{\textbf{Attribute-level}$\uparrow$ } & \multicolumn{2}{c}{\textbf{Object-level}$\uparrow$ } & \multicolumn{2}{c|}{\textbf{Attribute-level}$\uparrow$ } & \multicolumn{2}{c}{\textbf{Object-level}$\uparrow$ } & \multicolumn{2}{c}{\textbf{Attribute-level}$\uparrow$ } \\
            & Existence & Count & Position & Color & Existence & Count & Position & Color & Existence & Count & Position & Color \\
      \midrule
      Greedy & 165.67 & 120.00 & 110.67 & 148.33 & 137.00 & 93.00 & 75.00 & 125.00 & 167.00 & 120.00 & 105.00 & 145.00 \\
      DoLa  & 170.00 & 120.00 & 106.67 & 150.67 & 137.00 & 90.00 & 75.33 & 122.67 & 167.00 & 125.00 & 110.00 & 147.67 \\
      OPERA & 165.00 & 115.67 & 104.00 & 145.00 & 140.67 & 92.33 & 73.00 & 125.00 & 167.00 & 122.33 & 100.00 & 145.00 \\
      VCD   & 175.33 & 130.33 & 115.00 & 155.00 & 142.00 & 95.33 & 71.33 & 129.00 & 171.33 & 125.00 & 107.33 & 150.00 \\
      HALC  & 167.67 & 121.33 & 106.67 & 150.67 & 140.00 & 92.67 & 71.33 & 122.67 & 167.00 & 120.33 & 108.67 & 145.00 \\
      \midrule
      \rowcolor{aliceblue} Ours & \textbf{180.00} & \textbf{132.67} & \textbf{121.33} & \textbf{160.00} & \textbf{147.33} & \textbf{98.67} & \textbf{78.67} & \textbf{133.00} & \textbf{175.00} & \textbf{130.00} & \textbf{110.67} & \textbf{155.00} \\
      \bottomrule
      \end{tabular}%
    \caption{Results on the subset of the MME benchmark for evaluating object-level and attribute-level VH, where the best performances within each setting are bolded. We randomly run it five times to obtain the average result, with the whole statistical results in Appendix.
    }

    \label{tab:main-3}%
  \end{table*}%

\begin{table*}[htbp]
    \centering
            \renewcommand{\arraystretch}{0.9}
    \tabcolsep=0.175cm
    \small
      \begin{tabular}{cl|ccc|ccc}
      \toprule
      \multirow{2}[2]{*}{\textbf{G.}} & \multicolumn{1}{l|}{\multirow{2}[2]{*}{\textbf{Settings}}} & \multicolumn{3}{c|}{\textbf{LLaVA-1.5}} & \multicolumn{3}{c}{\textbf{MiniGPT-4}} \\
            &       & CHAIR$_i\downarrow$ & CHAIR$_s\downarrow$ &  TPS$\uparrow$  & CHAIR$_i\downarrow$ & CHAIR$_s\downarrow$ & TPS$\uparrow$ \\
      \midrule
      \multirow{2}[2]{*}{1} & w/o Whole Visual-Aware Token Selection (i.e., Eq.~\ref{eq:select}) & 6.43  & 19.75 & 25.54  & 11.63 & 30.51 & 27.55 \\
            & w/o Visual Perception Score $P$ in Eq.~\ref{eq:select} & 6.06  & 19.20  & 27.80  & 11.57 & 31.05 & 30.96 \\
      \midrule
      \multirow{2}[2]{*}{2} & w/o Whole SVCD (i.e., Eq.~\ref{eq:sparse-con}) & 6.91  & 21.42 & \textbf{30.68 } & 11.85 & 30.93 & \textbf{35.83} \\
            & w/o Mask-based Sparsification $S^{m}$ in Eq.~\ref{eq:sparse-con}& 6.31  & 18.85 & 27.47  & 11.58 & 31.26 & 30.30  \\
      \midrule
      3     & w/o Sinking Attention Penalty (i.e., Eq.~\ref{eq:penilty}) & 6.32  & 19.39 & 27.96  & 11.52 & 31.04 & 30.92 \\
      \midrule
      \rowcolor{aliceblue} 4     & Our Full VASparse & \textbf{5.82} & \textbf{18.51} & 27.73  & \textbf{11.35} & \textbf{30.19} & 30.87 \\
      \bottomrule
      \end{tabular}%
    \caption{ Ablation experiments on the CHAIR benchmark, with the best results highlighted in bold and the whole results in Appendix.}
    \vspace{-0.3em}
    \label{tab:ablation}%
  \end{table*}%

  \begin{table}[htbp]
    \centering
            \renewcommand{\arraystretch}{0.9}
    \tabcolsep=0.25cm
    \small
      \begin{tabular}{l|ccc}
      \toprule
           \textbf{Methods} & \textbf{LLaVA-1.5} & \textbf{MiniGPT-4} & \textbf{mPLUG-Owl2} \\
      \midrule
      Greedy & 36.3  & 46.7  & 42.3 \\
      OPERA & 34.2  & 45.9  & 41.7 \\
      VCD   & 34.6  & 46.0  & 41.9 \\
      HALC  & 33.9  & 45.8  & 41.7 \\
      \rowcolor{aliceblue} Ours  & \textbf{33.5} & \textbf{45.2} & \textbf{41.1} \\
      \bottomrule
      \end{tabular}%
    \caption{Performance (SHR) comparison on GPT-4 assisted benchmark, where, the lower value denotes the lower VH. }
    \label{tab:gpt-4}%
\vspace{-0.3em}
    
  \end{table}%

\textbf{Benchmarks.}~
Following common settings~\cite{Leng2023MitigatingOH, Chen2024HALCOH,Yin2023WoodpeckerHC}, 
We evaluate the effectiveness of our VASparse in VH mitigation on four popular benchmarks:
(1) quantitative metrics CHAIR~\cite{Rohrbach2018ObjectHI} on MSCOCO dataset~\cite{Lin2014MicrosoftCC};
(2) the offline Polling-based Object Probing Evaluation (POPE)~\cite{Li2023EvaluatingOH,Chen2024HALCOH} on the MSCOCO dataset;
(3) general-purposed Multimodal Large Language Model Evaluation (MME) benchmark~\cite{Fu2023MMEAC};
(4) GPT-4 assisted benchmark~\cite{zhao2023beyond} relies on the advanced GPT-4 to judge the fine-grained VH and calculate
Sentence-level Hallucination Ratio (SHR). 

\textbf{Baselines.}~
We compare our VASparse with greedy decoding and beam search decoding, and various state-of-the-art (SOTA) decoding methods as baselines,
including DoLa~\cite{Chuang2023DoLaDB}, OPERA~\cite{Huang2023OPERAAH}, VCD~\cite{Leng2023MitigatingOH}, SID~\cite{huo2024selfintrospectivedecodingalleviatinghallucinations} and HALC~\cite{Chen2024HALCOH}.
We also compare the post-processing VH elimination method (i.e., Woodpecker~\cite{Yin2023WoodpeckerHC}, LURE~\cite{Zhou2023AnalyzingAM}) with some token sparsity methods (i.e., FastV~\cite{chen2024image} and SparseVLMs~\cite{zhang2024sparsevlm}).

\textbf{Backbones.}~
Following previous settings~\cite{Leng2023MitigatingOH, Chen2024HALCOH}, 
we select popular LVLMs families, e.g., LLaVA-1.5~\cite{Liu2023VisualIT}, MiniGPT-4~\cite{Chen2023MiniGPTv2LL} and mPLUG-Owl2~\cite{Ye2023mPLUGOwl2RM} as the base modal for
all baselines except Woodpecker and LURE, where, Woodpecker and LURE utilize extra LLMs, i.e., ChatGPT~\cite{Brown2020LanguageMA} and GPT-4~\cite{Achiam2023GPT4TR}, for self-correction and distillation.
We investigate the VH of these LVLMs under different decoding to evaluate the effectiveness of our VASparse.

\textbf{Settings.}~
We implement the proposed VASparse based on HuggingFace Transformers~\cite{Wolf2019HuggingFacesTS} and combine it with beam search for decoding.
We evaluate settings with maximum generation lengths $L_{max}$ of 64 and 512. 
When $L_{max}$ is 64, the beam size is set to 3, and for $L_{max}=512$, it is set to 2. 
The sparsity rate top-$S$ is set to 0.9 times $L$, and the image masking sparsity rate for $S^{m}$ is set to 0.5. 
The hyperparameter $\lambda$ in Eq.~\ref{eq:select}, $\alpha$ in Eq.~\ref{eq:sparse-con} and $\beta$ in Section~\ref{subsec:attn} are set to 0.1. 
The decoding process of LVLMs and all experiments are performed on 8 A100 GPUs.
For token sparsity methods, we retain 75\% of tokens during inference.
Other methods use the settings as described in original papers. 
More details and results under $L_{max}=512$ are provided in Appendix.
\subsection{Main Results}

\noindent\textbf{CHAIR Evaluation.}~
Following HALC~\cite{Chen2024HALCOH}, we set `\textit{Please describe this image in detail.}' as the input prompt
and utilize generated tokens per second (TPS) to evaluate the efficiency, as results are shown in Table~\ref{tab:main-1}.
Based on the results, we have several detailed observations:
(1) It can be observed that our method significantly outperforms existing decoding and post-processing baselines for reducing VH. 
Our VASparse achieved the lowest VH rate at both the sentence and instance levels across three families of LVLMs,
which demonstrates the superiority and generalizability of our method in alleviating VH.
(2) Compared to SOTA decoding methods, VASparse maintains competitive decoding speed without secondary decoding or reprocessing via extra LLMs,
e.g., achieving speeds that are 12.9$\times$ and 6.4$\times$ faster than HALC~\cite{Chen2024HALCOH} and OPERA~\cite{Huang2023OPERAAH}, respectively.
(3) Although the sparsification method accelerates the inference speed, it exacerbates visual ambiguity, which in turn aggravates VH.

\noindent\textbf{POPE Evaluation.}~
Following HALC~\cite{Chen2024HALCOH}, we utilize offline POPE (OPOPE) benchmark with F1-score as metrics to evaluate VH, which replaces the live interactions of POPE with offline checks.
As shown in Table~\ref{tab:main-2}, we have several observations: 
(1) VASparse consistently achieves optimal results in most settings, outperforming both SOTA decoding methods and post-processing methods. This further demonstrates the effectiveness of VASparse;
(2) VASparse effectively mitigates VH across three different LVLM architectures, demonstrating the versatility and plug-and-play nature.

\noindent\textbf{MME Benchmarks.}~
Following~\cite{Yin2023WoodpeckerHC, Leng2023MitigatingOH,Chen2024HALCOH},
we adopt object-level subsets (``existence'' and ``count'') and attribute-level subsets ( ``position'' and ``color'') of MME benchmark~\cite{Fu2023MMEAC}.
to evaluate VH.
As shown in Table~\ref{tab:main-3}, we can observe that:
(1) Our VASparse can significantly reduce object and attribute hallucination, and achieve optimal VH mitigation performance.
(2) HALC and OPERA do not exhibit significant VH mitigation on the MME benchmark.
This is because the MME evaluation is designed as a binary classification task, requiring LVLMs to output only a few tokens, which limits the effectiveness of methods that need to decode sequences of a certain length
and handle special entity tokens.

\noindent\textbf{GPT-4 Assisted Benchmarks.}~
We conduct experiments on the GPT-4 assisted benchmark to evaluate the fine-grained VH of different methods, and the results are presented in Table~\ref{tab:gpt-4}. 
We can observe that our VASparse achieved the best SHR metric among the four LVLMs, which further confirms the superiority of our method in mitigating VH.

\subsection{Method Analysis}
We conduct ablation experiments using CHAIR on MSCOCO to evaluate the effectiveness of the components of our proposed VASparse in detail.
Specifically, we evaluate the effectiveness of the components by removing or modifying the specific settings as results shown in Table~\ref{tab:gpt-4}.

\noindent\textbf{Effect of the Visual-Aware Token Selection.}~
As shown in Groups 1 and 4 in Table~\ref{tab:ablation}, 
removing the whole visual-aware token selection strategy leads to a performance decrease and reduces decoding speed.
This suggests that sparsifying the model's decoding sequence to some extent can mitigate the language bias in LVLMs and reduce the involvement of certain tokens in attention computation.
Moreover, removing the visual perception score also results in a performance decline.
These results consistently demonstrate the effectiveness of our visual-aware token selection strategy.

\noindent\textbf{Effect of the Sparse-based Visual Contrastive Decoding.}~
To evaluate the effectiveness of our sparse-based visual contrastive decoding (SVCD), we remove both the full SVCD and the mask-based sparsification $S^{m}$ in Eq.~\ref{eq:sparse-con}.
As shown in Groups 2 and 4 of Table~\ref{tab:ablation}, we observe a significant performance decline, which further validates the effectiveness of our SVCD and mask-based sparsification strategy.

\noindent\textbf{Effect of the Sinking Attention Calibration.}~
Moreover, we removed the calibration mechanism for the sinking attention, and observed a further decline in the method's VH mitigation effect. 
This further demonstrates the relevance of sinking attention to VH and the effectiveness of the proposed attention calibration strategy.

\begin{figure}
    \centering
    \begin{subfigure}{0.49\linewidth}
      \includegraphics[width=1\linewidth]{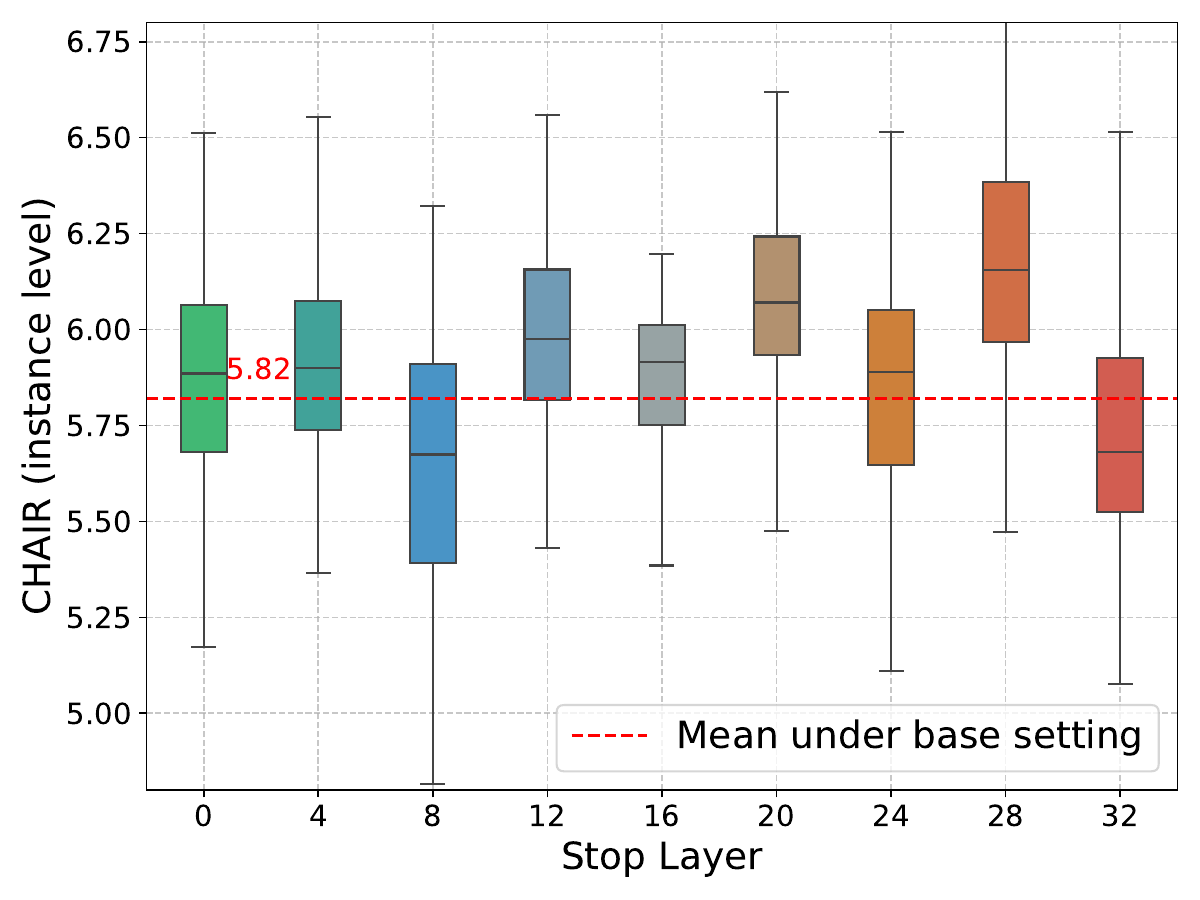}
      \caption{CHAIR$_i$ evaluation results.}
      \label{fig:exp-1}
    \end{subfigure}
    \hfill
    \begin{subfigure}{0.49\linewidth}
      \includegraphics[width=1\linewidth]{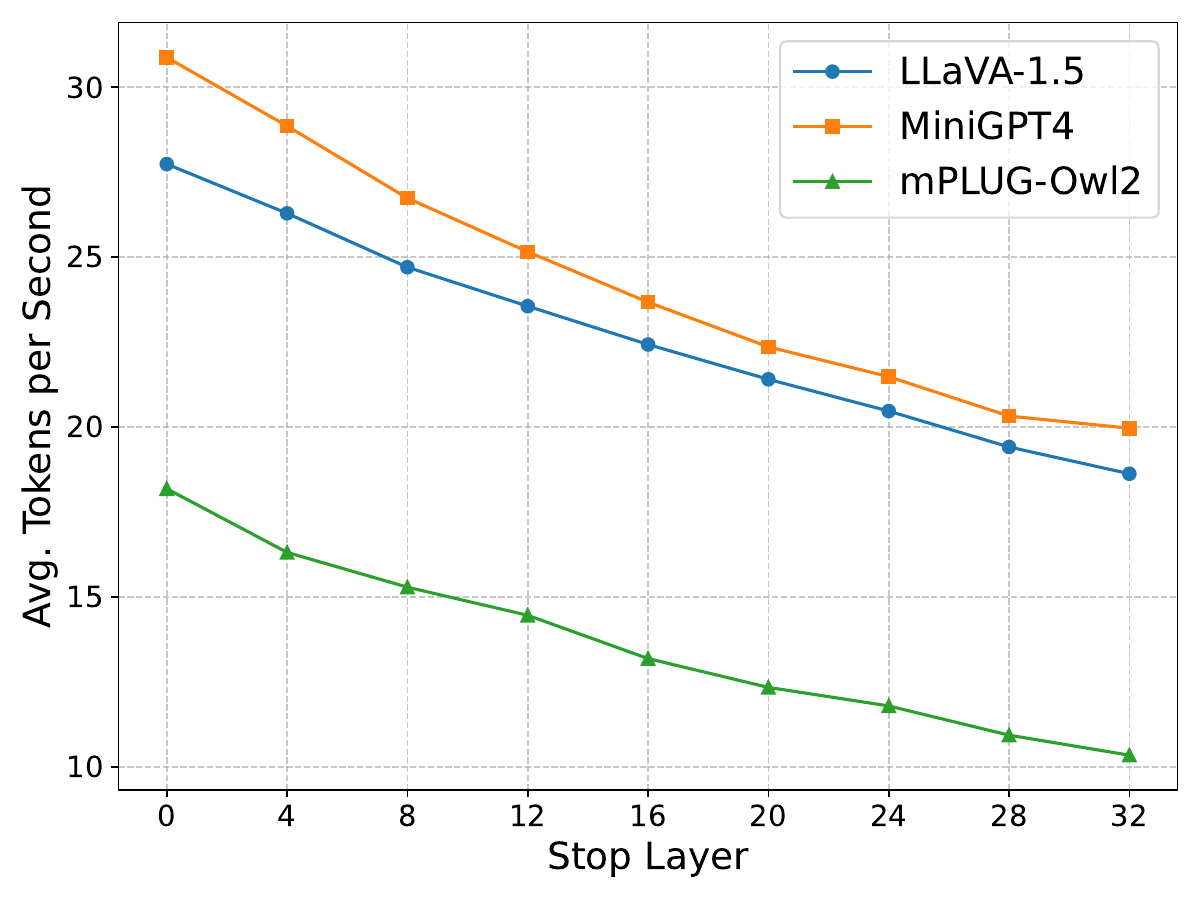}
      \caption{TPS during decoding.}
      \label{fig:exp-2}
    \end{subfigure}
    \caption{Performance and efficiency analysis of different logit sources: (a) the impact of using different early stopping layers on LLaVA-1.5 performance; (b) the impact of using different early stopping layers on decoding speeds (TPS).}
        \label{fig:exp}
\vspace{-0.3em}
        
  \end{figure}

\noindent\textbf{Decoding Efficiency Analysis.}~
To further validate the effect of using embedding features to compute the proposed SVCD, we calculate the contrastive logits from features at different depths of the LVLM decoder to calibrate the distribution, and observe performance and decoding speed, as shown in Figure~\ref{fig:exp}.
We observe that by using only embedded features (i.e., stop layer is 0), our method already achieves good VH mitigation performance while attaining optimal decoding speed.
In this way, our VASparse effectively avoids the time-consuming secondary decoding process, achieving a balance between performance and efficiency.

\section{Conclusion}
\label{sec:conclusion}

This work proposes an efficient, plug-and-play decoding strategy, VASparse, to mitigate VH in LVLMs. 
Inspired by the sparse activation pattern of LVLMs and the role of visual-agnostic token sparsification in worsening VH, we propose a visual-aware token selection strategy during decoding. Subsequently, we innovatively introduce sparse-based visual contrastive decoding to recalibrate the logits without secondary decoding, and adjust sinking attention.
Extensive experiments show the effectiveness of VASparse in reducing VH across various benchmarks and LVLM families.

\section*{Acknowledgements}
This work is supported by Guangdong Provincial Key Laboratory of Ultra High Definition Immersive Media Technology(Grant No. 2024B1212010006)

{
    \small
    \bibliographystyle{ieeenat_fullname}
    \bibliography{main}
}
\maketitlesupplementary

\section{Experimental Detials}~
\subsection{Experimental Setting}
For the settings of the CHAIR and POPE benchmarks, we evaluated the results with the maximum generation token limits of LVLM $L_{max}$ set to 64 and 512, respectively. 
For the GPT4-assist benchmark~\cite{zhao2023beyond}, we referred to prior work and adopted SHR Evaluation. For the GPT-4 settings, we followed the GPT4-assist configurations and used OpenAI's \texttt{gpt-4-0613} version API for evaluation.
The parameters for LVLMs were set as follows: \texttt{Top-k=False}, \texttt{Top-p=1}, and \texttt{Temperature=1}.
All our experiments, including decoding speed statistics, are conducted on Tesla A100-80G GPUs.

For the proposed token selection strategy, we do not perform sparsification at every decoding step, as this would result in excessive sparsification at each step, leading to overly short generated sequences. In practice, we perform sparsification only after decoding a certain length of new tokens, denoted as $L_s$.
For $L_{\text{max}} = 64$, the beam size is set to 3, and $L_s$ is set to 32.  
For $L_{\text{max}} = 512$, the beam size is set to 2, and $L_s$ is set to 16.
Additionally, in our method, the adaptive plausibility threshold is set to 0.1.

Regarding the comparison of VASparse with SOTAs that are specifically designed for VH mitigation, we adopt the code, hyper-parameters, and pre-trained models of each method outlined in their public repositories and papers respectively.
Specifically, for DoLa~\cite{Chuang2023DoLaDB}, the parameters are set as follows: the repetition penalty is 1.2, the adaptive plausibility threshold is 0.1, and the pre-mature layers are $[0, 2, \ldots, 32]$. For beam search-based OPERA~\cite{Huang2023OPERAAH} hyperparameters are set as follows: the self-attention weights scale factor is 50, the attending retrospection threshold is 15, the beam size is 3, and the penalty weights are 1.
The VCD~\cite{Leng2023MitigatingOH} hyperparameters are set as follows: the amplification factor is 1, the adaptive plausibility threshold is 0.1, and the diffusion noise step is 500.
The HALC~\cite{Chen2024HALCOH} hyperparameters are set as follows: the amplification factor is 0.05, the JSD buffer size is 6, the beam size is 1, the FOV sampling uses exponential expansion, the number of sampled FOVs is 4, the exponential growth factor is 0.6, and the adaptive plausibility threshold is 0.1.
For post-processing methods, such as LURE and Woodpecker, we follow the settings in HALC~\cite{Chen2024HALCOH}.
For the SID method~\cite{huo2024selfintrospectivedecodingalleviatinghallucinations}, we referred to the original configuration in their paper.
For all baselines, we follow their implementations and default hyper-parameters as reported in the paper.

\subsection{Setting of Empirical Studies}
In Section 3, we provide our empirical observations, where all empirical studies are based on LLaVA-1.5~\cite{Zhang2023VideoLLaMAAI}. For the hallucination evaluation results, experiments are conducted on 500 samples randomly selected from the MSCOCO dataset.
For decoding speed, we calculate the average number of tokens decoded per second by the model on the 500 samples.
Token sparsification methods, such as FastV~\cite{chen2024image} and SparseVLM~\cite{zhang2024sparsevlm}, directly prune image tokens.

\section{Proof of Theorem 1}

We aim to prove that in the following optimization problem, our strategy achieves a globally optimal solution:

\begin{equation}
\begin{aligned}
\min_{M} \quad & \mathcal{E}(M) = \sum_{i=1}^L   \left( y_i - M_i y_i \right)^2 - \lambda P_i M_i \\
\text{s.t.} \quad & M_i \in \{0, 1\}, \quad \forall i = 1, 2, \dots, L, \\
& \sum_{i=1}^L M_i = S,
\end{aligned}
\end{equation}

where:
\begin{itemize}
    \item $y_i = \langle q, K_i \rangle$ is the inner product of the query vector $q$ and the key matrix vector $K_i$.
    \item $P_i \geq 0$ is the selection probability, indicating the priority of selecting a specific element.
    \item $M_i \in \{0, 1\}$ denotes whether the $i$-th element is selected.
    \item The constraint requires exactly $S$ elements in $M$ to be 1.
\end{itemize}
The goal is to minimize the total error $\mathcal{E}(M)$ when selecting $S$ elements.

\textbf{Proof}
First, expand and simplify the objective function $\mathcal{E}(M)$:
\begin{equation}
\begin{aligned}
\mathcal{E}(M) &= \sum_{i=1}^L \left[   \left( y_i - M_i y_i \right)^2 - \lambda P_i M_i \right] \\
&= \sum_{i=1}^L \left[   y_i^2 (1 - M_i)^2 - \lambda P_i M_i \right].
\end{aligned}
\end{equation}

Since $M_i \in \{0, 1\}$, it follows that $M_i^2 = M_i$ and $(1 - M_i)^2 = 1 - 2M_i + M_i^2 = 1 - 2M_i + M_i$. Substituting these simplifications, we get:
\begin{equation}
\begin{aligned}
\mathcal{E}(M) &= \sum_{i=1}^L \left[   y_i^2 (1 - 2M_i + M_i) - \lambda P_i M_i \right] \\
&= \sum_{i=1}^L \left[   y_i^2 (1 - M_i) - \lambda P_i M_i \right].
\end{aligned}
\end{equation}

Next, remove the constant term $\sum_{i=1}^L   y_i^2$, as it does not affect the optimization:
\begin{equation}
\begin{aligned}
\mathcal{E}(M) &= \sum_{i=1}^L \left[   y_i^2 -   y_i^2 M_i - \lambda P_i M_i \right] \\
&= \sum_{i=1}^L \left[   y_i^2 - M_i (  y_i^2 + \lambda P_i) \right].
\end{aligned}
\end{equation}

Thus, the optimization problem can be equivalently transformed into maximizing the following objective function:
\begin{equation}
\begin{aligned}
\max_{M} \quad & \sum_{i=1}^L M_i (  y_i^2 + \lambda P_i) \\
\text{s.t.} \quad & M_i \in \{0, 1\}, \quad \forall i, \\
& \sum_{i=1}^L M_i = S.
\end{aligned}
\end{equation}

Our goal is now to select $S$ elements to maximize the total reward $\sum_{i=1}^L M_i \delta_i$, where:
\begin{equation}
\delta_i =   y_i^2 + \lambda P_i.
\end{equation}

\textbf{Characteristics of the Objective Function}
\begin{itemize}
    \item \textbf{Linearity:} The objective function is linear with respect to $M_i$, with no interaction terms between $M_i$ and $M_j$.
    \item \textbf{Independence:} The contribution of each $M_i$ to the total reward depends solely on its own $\delta_i$, independent of other variables $M_j$.
\end{itemize}

We employ the following selection strategy:
\begin{enumerate}
    \item Compute the marginal reward $\delta_i$ for each element:
    \begin{equation}
    \delta_i =   y_i^2 + \lambda P_i.
    \end{equation}
    \item Sort all elements by $\delta_i$ in descending order.
    \item Select the top $S$ elements, setting their corresponding $M_i$ to 1, and the rest to 0.
\end{enumerate}

\textbf{Optimality of the Strategy}
For any feasible solution $M$, we have:
\begin{equation}
\sum_{i=1}^L M_i = S, \quad M_i \in \{0,1\}.
\end{equation}
Define the total reward for a solution $M$ as:
\begin{equation}
R(M) = \sum_{i=1}^L M_i \delta_i.
\end{equation}
Let the solution chosen by our strategy be $M^{\text{ours}}$, with total reward:
\begin{equation}
R_{\text{ours}} = \sum_{i=1}^L M_i^{\text{ours}} \delta_i,
\end{equation}
where $M_i^{\text{ours}} = 1$ if $i$ belongs to the top $S$ elements with the highest $\delta_i$, and $M_i^{\text{ours}} = 0$ otherwise. Since $\delta_i$ is sorted in descending order, the elements chosen by our strategy have the highest individual scores.

For any element $i$ in $M$ such that $M_i = 1$, if its score $\delta_i$ is smaller than that of an unselected element $j$ (i.e., $M_j = 0$), swapping these two elements would result in a new total reward:
\begin{equation}
R'(M) = R(M) - \delta_i + \delta_j.
\end{equation}
Since $\delta_j > \delta_i$, this increases the total reward. Thus, any feasible solution $M$ with lower-scoring elements can always be improved by following our selection strategy.

Finally, for any feasible solution $M$, we have:
\begin{equation}
R_{\text{ours}} = \sum_{i=1}^L M_i^{\text{ours}} \delta_i \geq \sum_{i=1}^L M_i \delta_i = R(M).
\end{equation}

\textbf{Conclusion}
The total reward achieved by our algorithm is no less than that of any other feasible solution. Therefore, the solution provided by our strategy is globally optimal.

\begin{figure*}[htbp]
    \centering
    \includegraphics[width=0.23\textwidth]{ 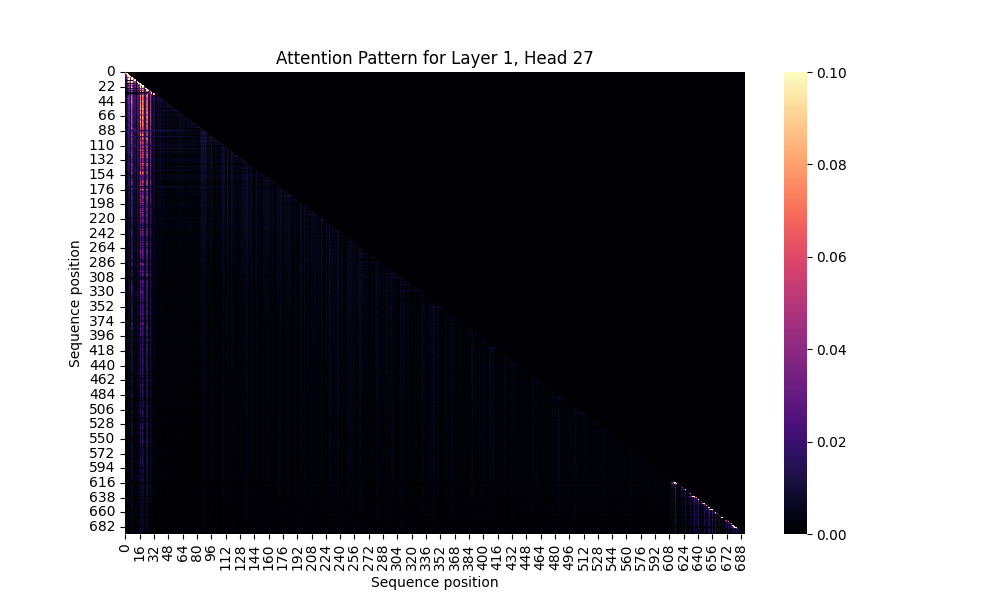}
    \hfill
    \includegraphics[width=0.23\textwidth]{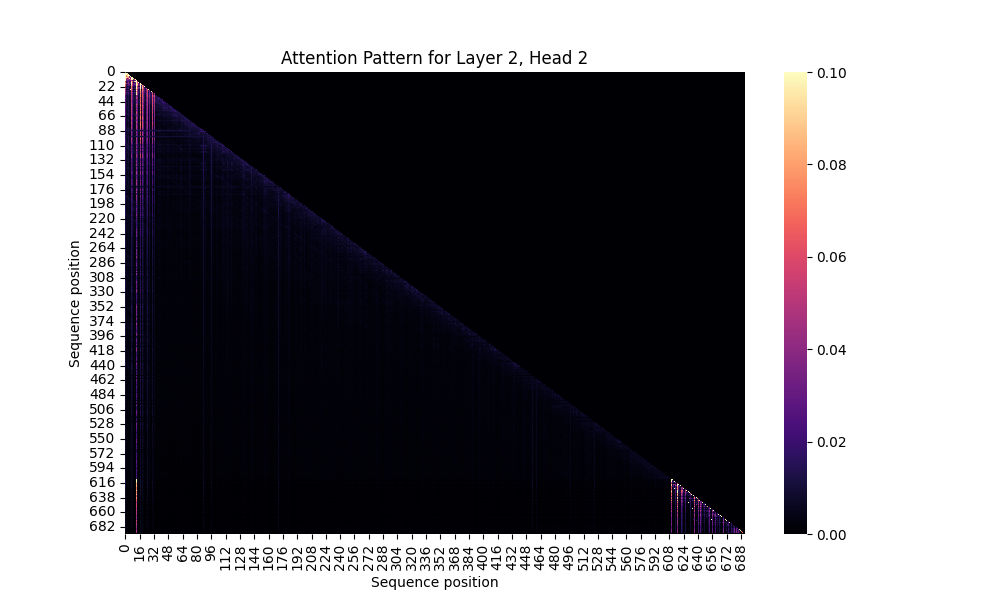}
    \hfill
    \includegraphics[width=0.23\textwidth]{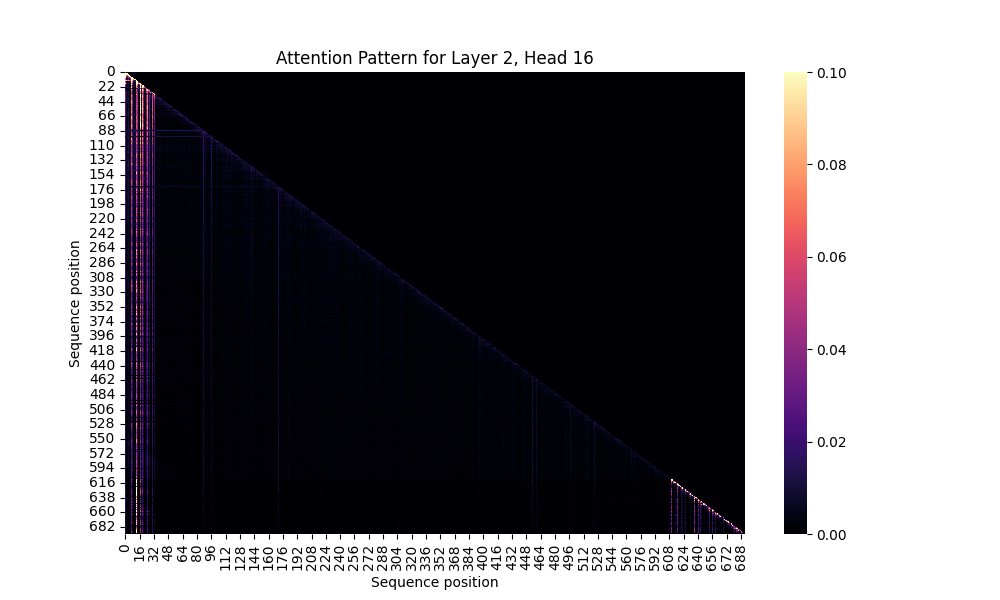}
    \hfill
    \includegraphics[width=0.23\textwidth]{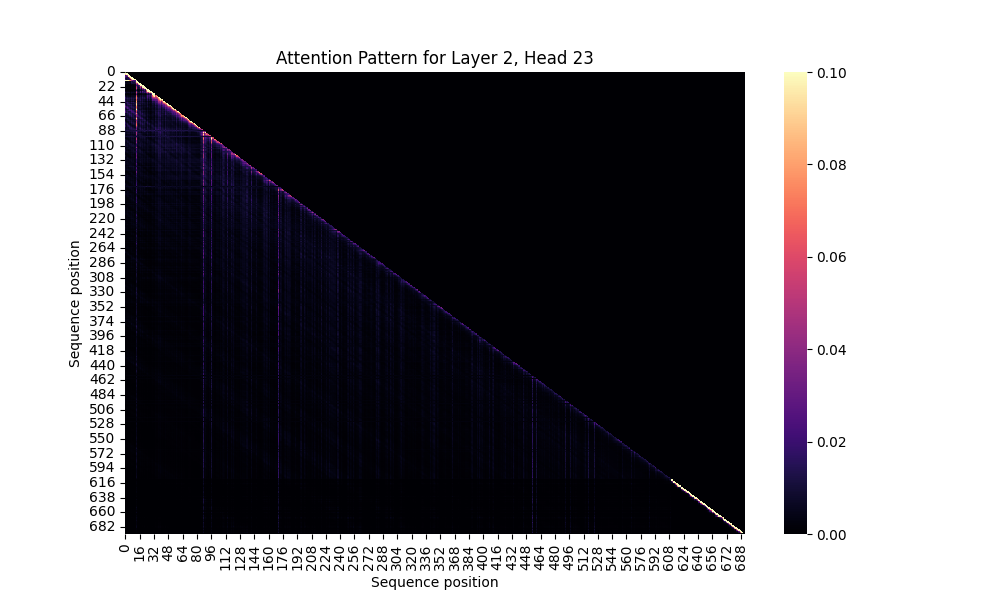}
    
    \vspace{0.5cm}
    
    \includegraphics[width=0.23\textwidth]{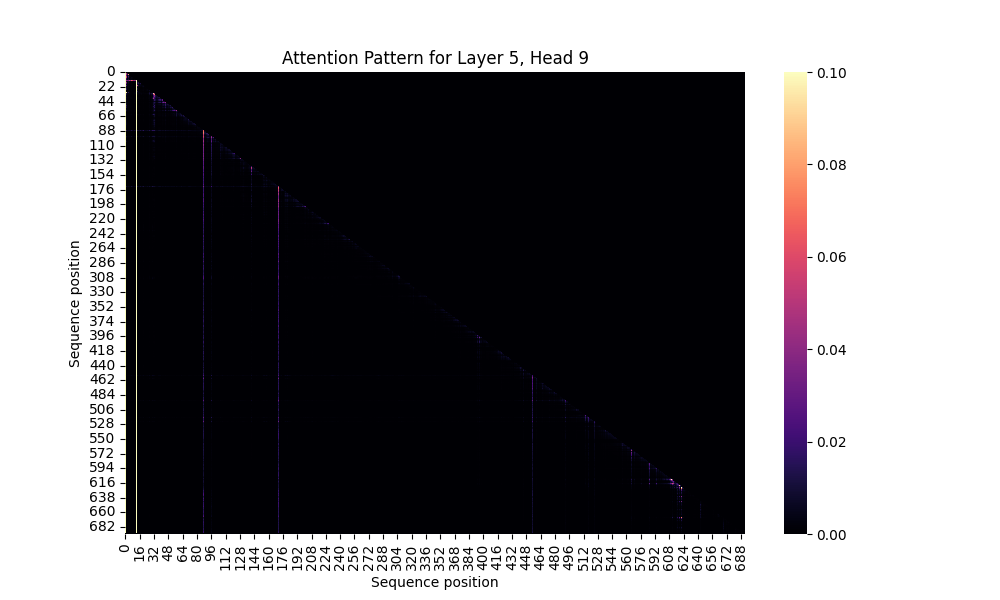}
    \hfill
    \includegraphics[width=0.23\textwidth]{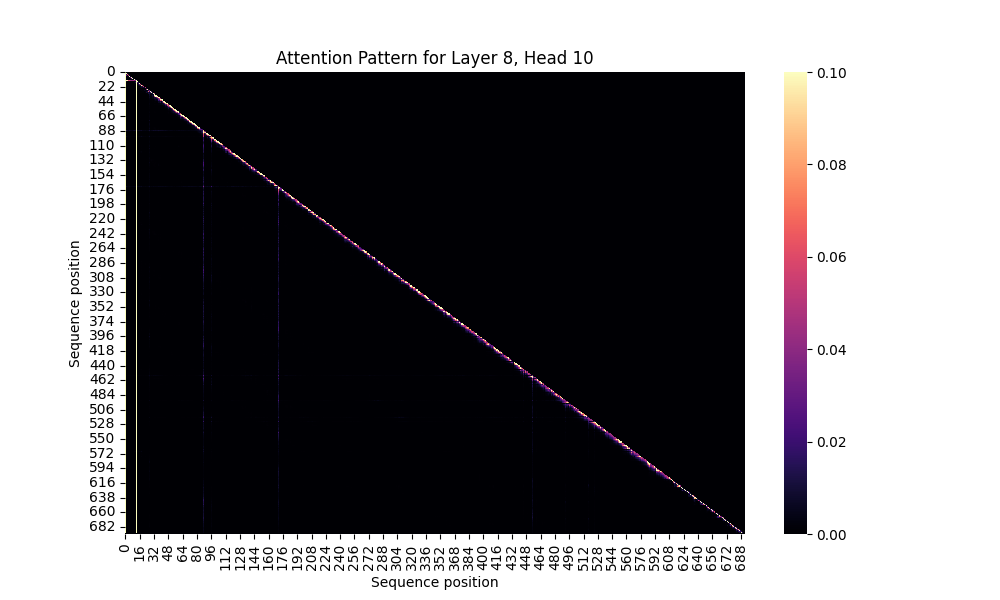}
    \hfill
    \includegraphics[width=0.23\textwidth]{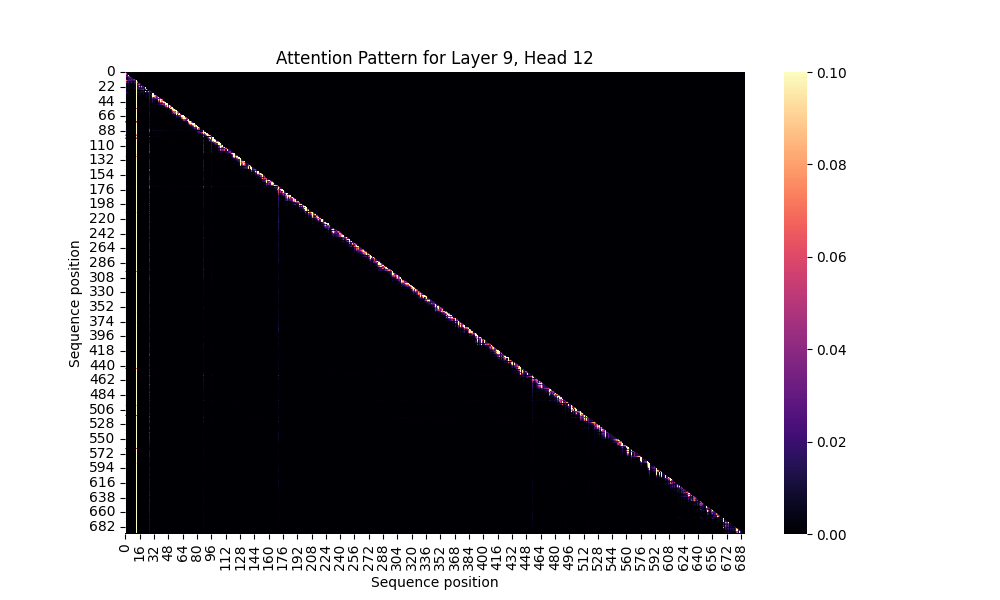}
    \hfill
    \includegraphics[width=0.23\textwidth]{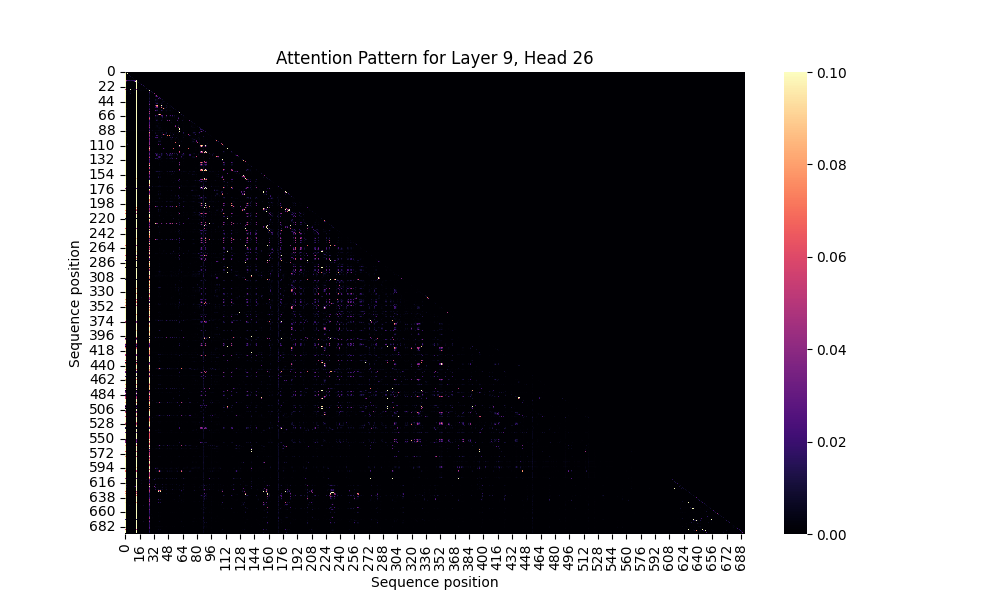}
    
    \vspace{0.5cm}
    
    \includegraphics[width=0.23\textwidth]{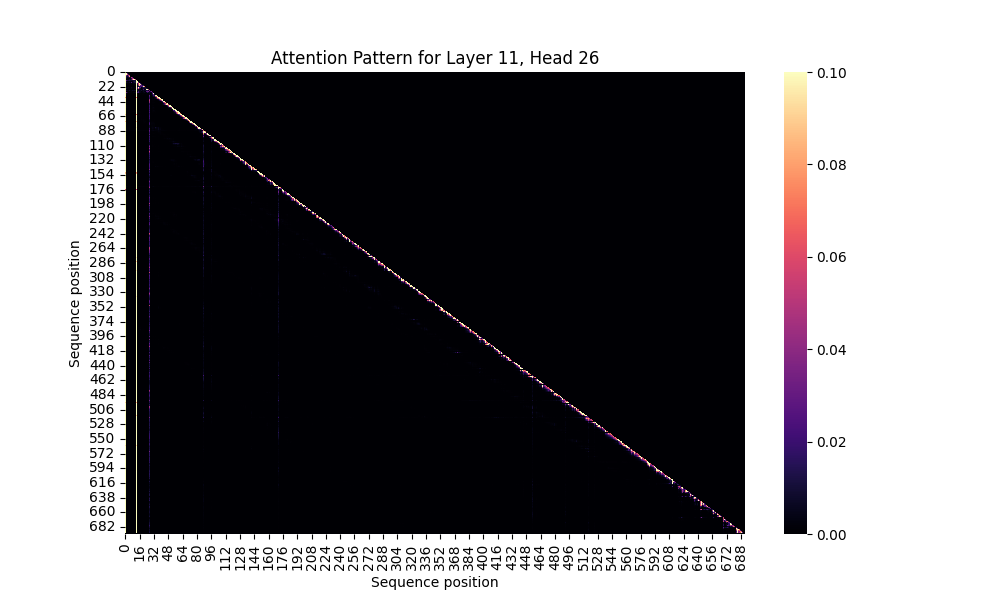}
    \hfill
    \includegraphics[width=0.23\textwidth]{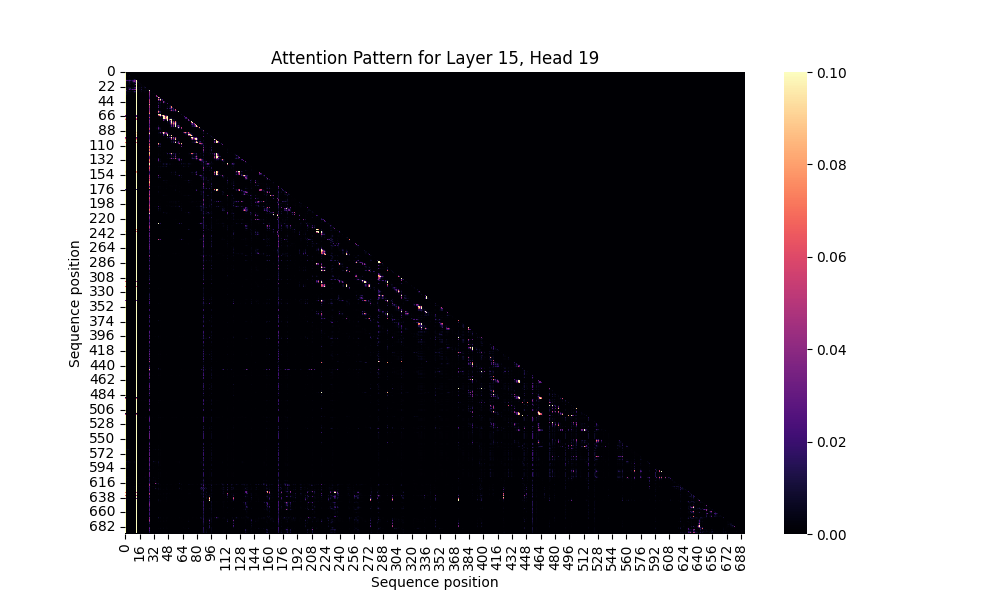}
    \hfill
    \includegraphics[width=0.23\textwidth]{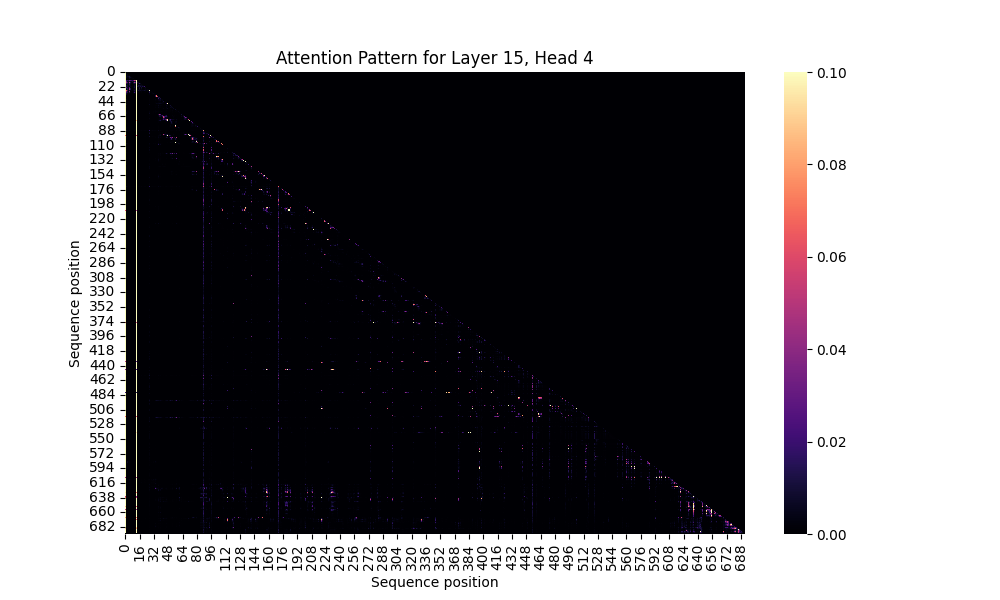}
    \hfill
    \includegraphics[width=0.23\textwidth]{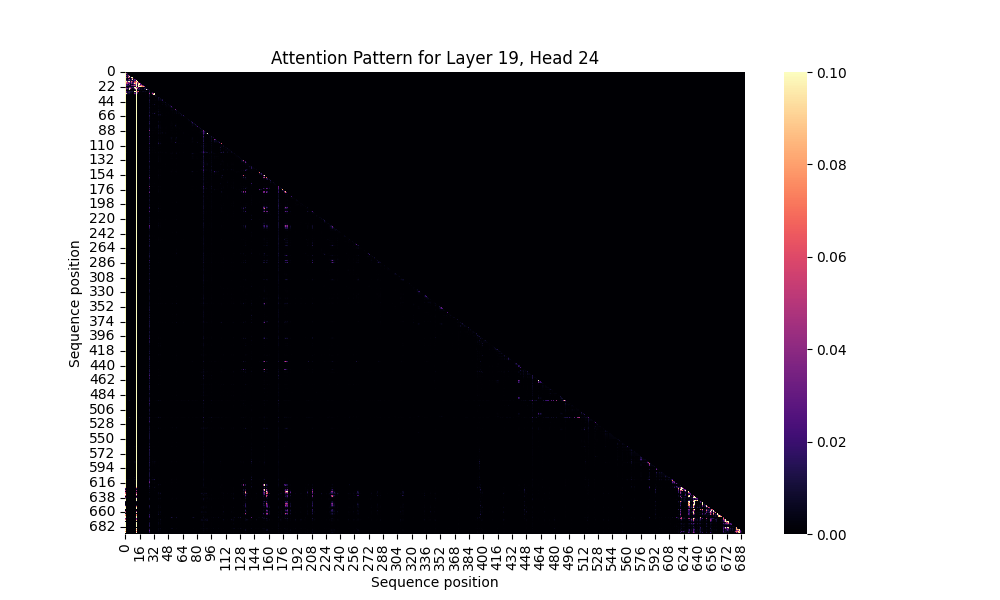}
    
    \vspace{0.5cm}
    
    \includegraphics[width=0.23\textwidth]{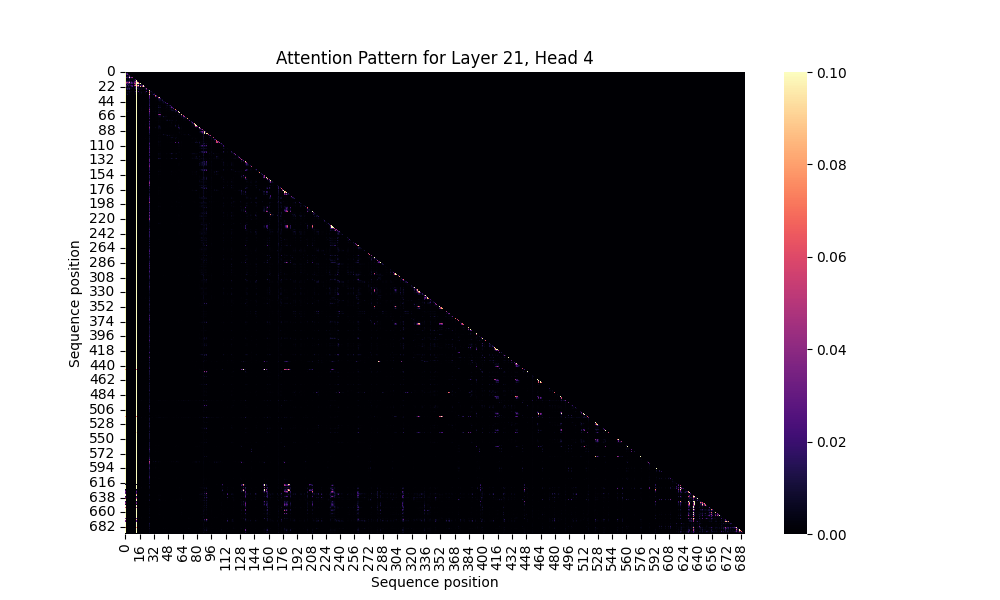}
    \hfill
    \includegraphics[width=0.23\textwidth]{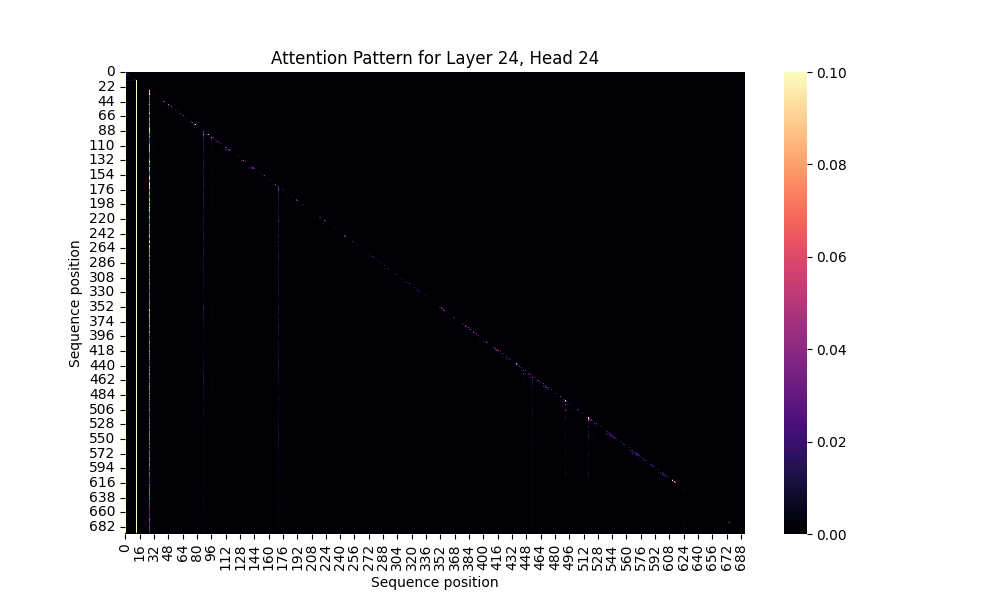}
    \hfill
    \includegraphics[width=0.23\textwidth]{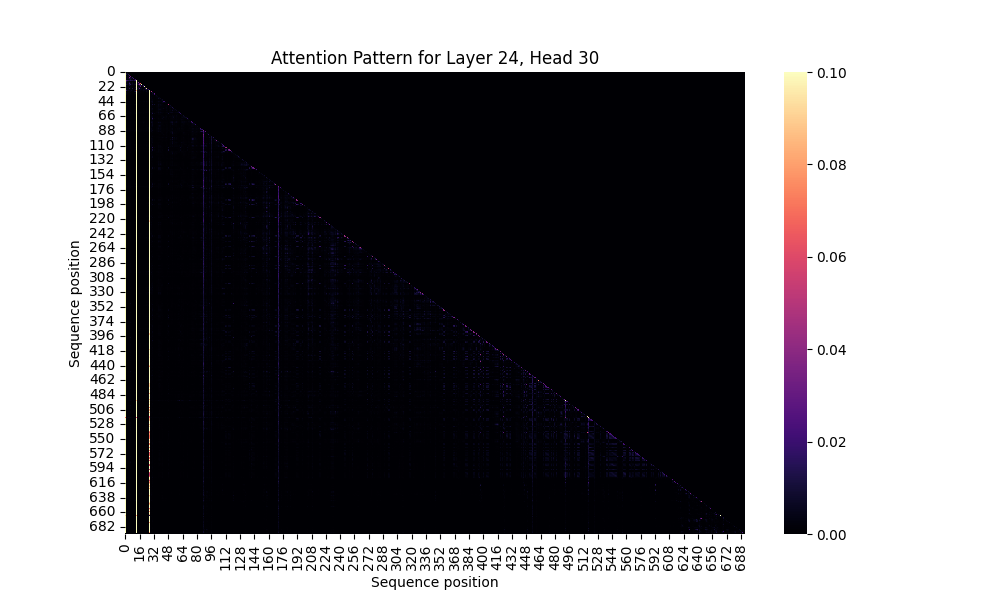}
    \hfill
    \includegraphics[width=0.23\textwidth]{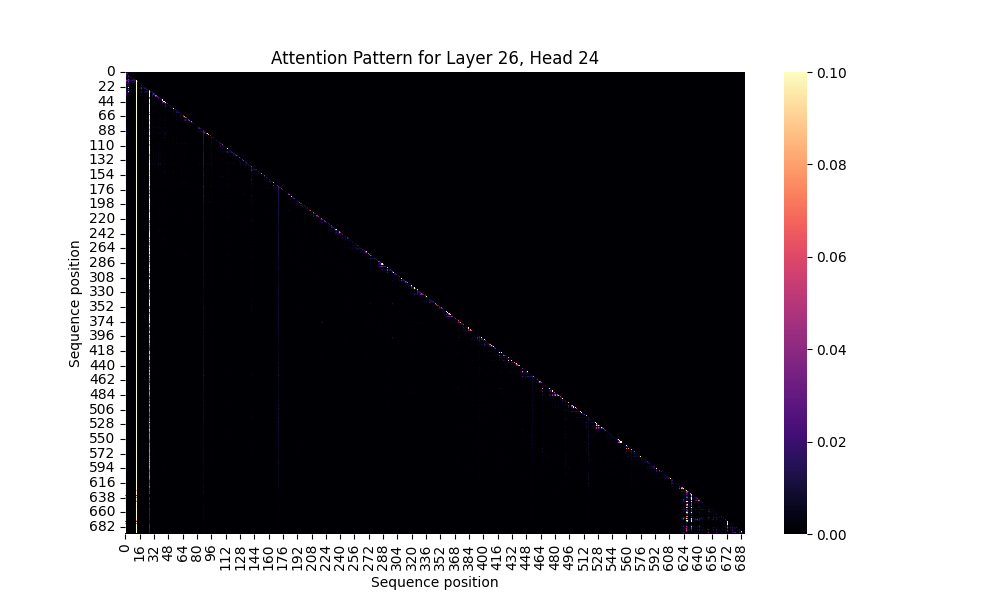}
    
    \vspace{0.5cm}
    
    \includegraphics[width=0.23\textwidth]{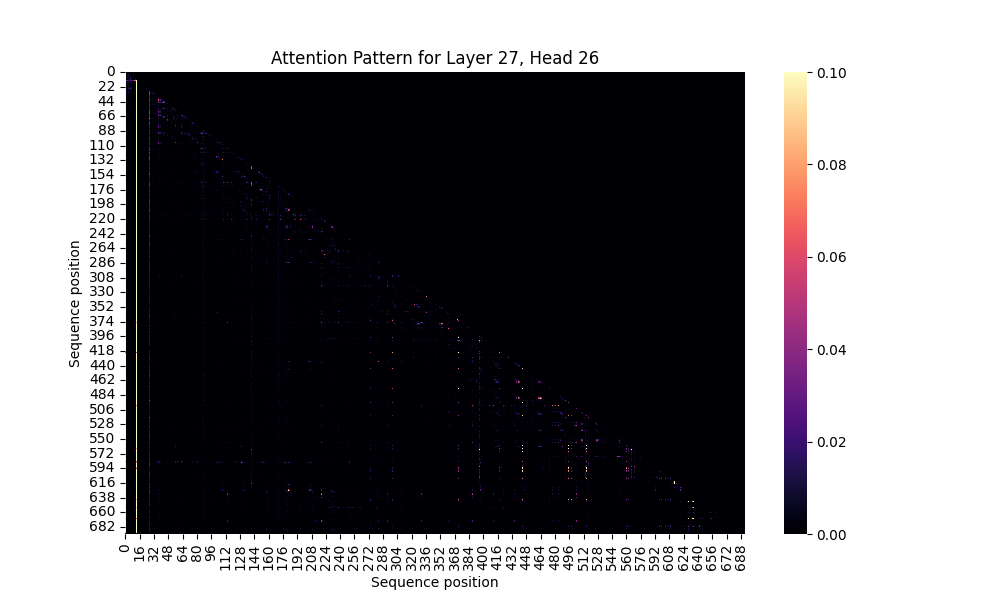}
    \hfill
    \includegraphics[width=0.23\textwidth]{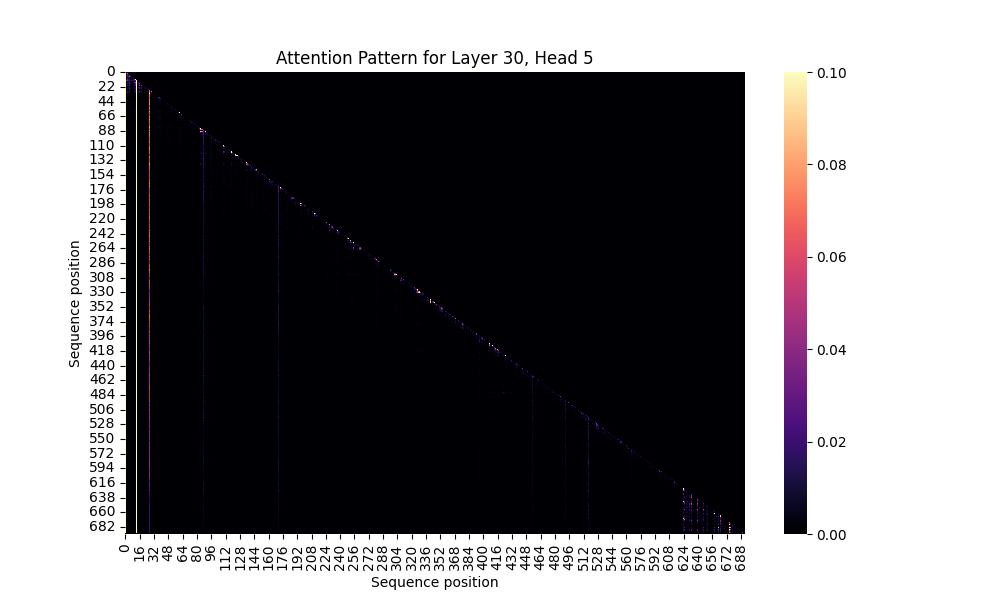}
    \hfill
    \includegraphics[width=0.23\textwidth]{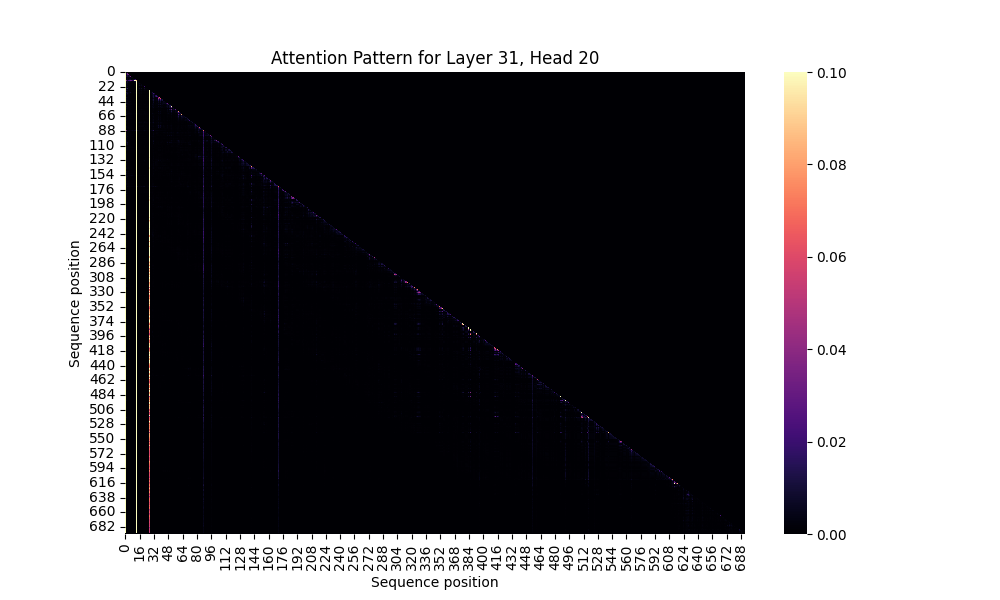}
    \hfill
    \includegraphics[width=0.23\textwidth]{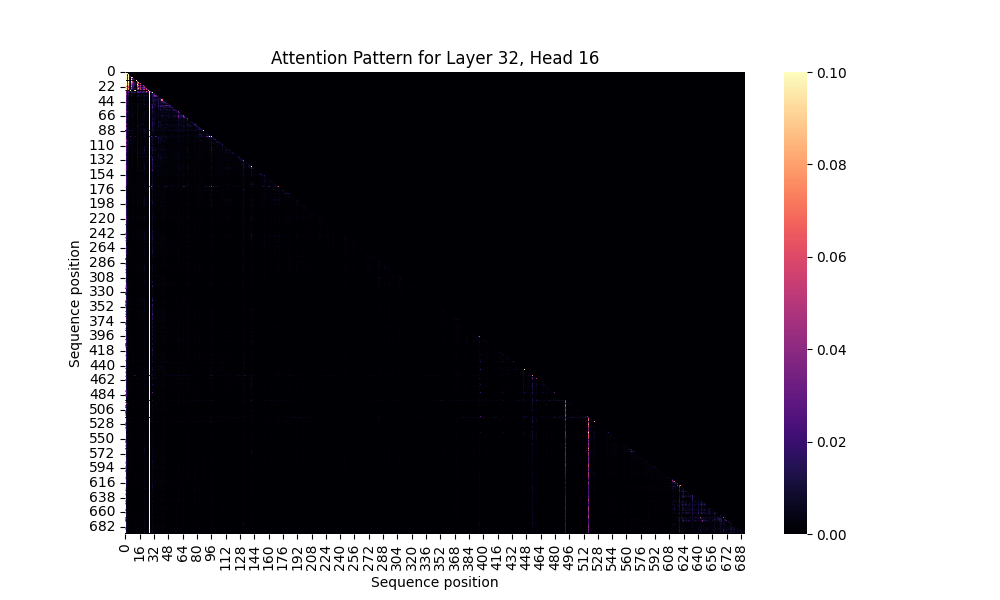}
    
    \caption{More visualization and evidence of sparsity of attention and sinking attention on the LLaVA-1.5.}
    \label{heads_simple}
\end{figure*}

\begin{table*}[htbp]
      \centering
        \begin{tabular}{c|ccc|ccc|ccc}
        \toprule
        \multirow{2}[2]{*}{Methods} & \multicolumn{3}{c|}{LLaVA-1.5} & \multicolumn{3}{c|}{MiniGPT-4} & \multicolumn{3}{c}{ mPLUG-Owl2} \\
              & CHAIR$_i\downarrow$ & CHAIR$_s\downarrow$ & TPS$\uparrow$   & CHAIR$_i\downarrow$ & CHAIR$_s\downarrow$ & TPS$\uparrow$    & CHAIR$_i\downarrow$ & CHAIR$_s\downarrow$ & TPS$\uparrow$  \\
        \midrule
        FastV$*$ & 17.62  & 51.94  & 33.18  & 19.67  & 54.59  & 38.05  & 22.65  & 67.68  & 24.37 \\
        SparseVLM$*$ & 18.09  & 52.40  & 32.06  & 19.85  & 55.27  & 37.49  & 23.04  & 68.42  & 23.19 \\
        Woodpecker$\dagger$ & 13.27  & 49.72  & -     & 13.76  & 44.07  & -     & 18.39  & 59.58  & - \\
        LURE$\dagger$  & 13.08  & 47.95  & -     & 13.49  & 43.92  & -     & 17.85  & 57.73  & - \\
        Greedy & 14.63  & 49.66  & 31.17  & 14.06  & 43.65  & 36.28  & 19.07  & 61.28  & 19.96 \\
        Beam Search & 13.62  & 48.89  & 29.89  & 13.90  & 44.45  & 32.10  & 17.12  & 54.66  & 19.58 \\
        OPERA & 12.98  & 47.60  & 4.07  & 15.42  & 42.42  & 5.27  & 17.86  & 56.29  & 3.49 \\
        VCD   & 14.82  & 49.76  & 17.55  & 17.09  & 43.80  & 17.68  & 19.46  & 62.44  & 9.77 \\
        DoLa  & 13.75  & 50.03  & 23.40  & 13.85  & 44.20  & 24.75  & 18.43  & 60.18  & 14.23 \\
        SID   & 13.29  & 47.09  & 19.57  & 13.68  & 43.65  & 22.67  & 18.47  & 60.82  & 12.85 \\
        HALC  & 12.93  & 46.35  & 2.04  & 13.73  & 43.68  & 3.68  & 17.63  & 56.12  & 1.50 \\
        Ours  & \textbf{12.46 } & \textbf{46.21 } & 27.53  & \textbf{13.29 } & \textbf{43.02 } & 29.74  & \textbf{17.02 } & \textbf{53.70 } & 17.86 \\
        \bottomrule
        \end{tabular}%
        \caption{Comparison of the average CHAIR evaluation results (instance levels CHAIR$_i$ and sentence levels CHAIR$_s$ )and token per second (TPS) during decoding with different baselines on MSCOCO datasets of five random runs. $^*$ represents the image token sparsity method and $\dagger$ is the post-hoc methods.}
      \label{tab:chair-Appendix}%
    \end{table*}%

    \begin{table*}[htbp]
          \centering
          \begin{tabular}{c|c|cccccc}
            \toprule
            \multirow{3}[2]{*}{} & \multirow{3}[2]{*}{Methods} & \multicolumn{6}{c}{Max New Token 512} \\
                  &       & \multicolumn{2}{c}{Random} & \multicolumn{2}{c}{Popular} & \multicolumn{2}{c}{Adversarial} \\
                  &       & Accuracy & F1 score & Accuracy & F1 score & Accuracy & F1 score \\
            \midrule
            \multirow{8}[2]{*}{LLaVA-1.5} & Greedy & 77.19  & 71.74  & 72.74  & 67.99  & 71.18  & 66.76  \\
                  & Beam Search & 78.38  & \textbf{73.60 } & 75.06  & 70.74  & 72.87  & \textbf{68.96 } \\
                  & OPERA & 78.01  & 72.98  & 74.31  & 69.81  & 73.25  & 68.95  \\
                  & VCD   & 77.82  & 72.98  & 74.56  & 70.19  & 72.62  & 68.63  \\
                  & DoLa  & 76.69  & 71.07  & 72.12  & 67.26  & 70.80  & 66.23  \\
                  & SID   & 77.93  & 72.84  & 74.89  & 69.34  & 72.77  & 68.30  \\
                  & HALC  & 77.08  & 72.16  & 74.15  & 69.09  & 72.46  & 68.04  \\
                  & Ours  & \textbf{78.57 } & 72.33  & \textbf{75.16 } & \textbf{70.51 } & \textbf{73.37 } & 68.88  \\
            \midrule
            \multirow{8}[2]{*}{MiniGPT4} & Greedy & 69.14  & 56.55  & 65.84  & 54.04  & 65.67  & 53.91  \\
                  & Beam Search & 68.90  & 55.78  & 65.67  & 53.32  & 65.61  & 53.28  \\
                  & OPERA & 69.77  & 57.24  & \textbf{66.90 } & 55.04  & 65.38  & 53.85  \\
                  & VCD   & 69.32  & 57.05  & 65.14  & 53.89  & 65.25  & 53.98  \\
                  & DoLa  & 69.02  & 56.31  & 66.08  & 54.07  & 65.84  & 53.90  \\
                  & SID   & 69.05  & 56.53  & 65.58  & 53.53  & 65.45  & 53.52  \\
                  & HALC  & 69.13  & 56.86  & 65.62  & 53.63  & 65.73  & 53.69  \\
                  & Ours  & \textbf{69.84 } & \textbf{57.36 } & 66.31  & \textbf{55.68 } & \textbf{66.02 } & \textbf{54.10 } \\
            \midrule
            \multirow{8}[2]{*}{ mPLUG-Owl2} & Greedy & 76.21  & 70.16  & 71.61  & 81.48  & 69.38  & 64.63  \\
                  & Beam Search & 75.83  & 69.87  & 71.83  & 81.75  & 69.02  & 64.29  \\
                  & OPERA & 73.56  & 65.33  & 70.32  & \textbf{84.43 } & 67.90  & 60.82  \\
                  & VCD   & 75.74  & 69.16  & 70.67  & 80.63  & 69.08  & 63.77  \\
                  & DoLa  & 76.33  & 70.22  & 71.67  & 81.72  & 69.55  & 64.71  \\
                  & SID   & 75.72  & 69.31  & 71.79  & 81.90  & 69.12  & 64.10  \\
                  & HALC  & 75.62  & 69.04  & 70.24  & 82.40  & 68.35  & 63.51  \\
                  & Ours  & \textbf{76.51}  & \textbf{70.45}  & \textbf{72.19}  & 82.44  & \textbf{69.72 } & \textbf{64.98 } \\
            \bottomrule
            \end{tabular}%
          \caption{Comparison of the average Accuracy and F1-score evaluation results under different settings (i.e., \textit{ Random, Popular, Adversarial}) with different baselines and our VASparse on offline POPE benchmark~\cite{Li2023EvaluatingOH, Chen2024HALCOH} of five random runs. Higher F1-score indicate better performance and bold indicates the best results.  We set the maximum generated length to 512.}
          \label{tab:opope-appendix-512}%
        \end{table*}%
        
    \begin{table*}[htbp]
      \centering
        \begin{tabular}{c|c|cccccc}
        \toprule
        \multirow{3}[2]{*}{} & \multirow{3}[2]{*}{Methods} & \multicolumn{6}{c}{Max New Token 64} \\
              &       & \multicolumn{2}{c}{Random} & \multicolumn{2}{c}{Popular} & \multicolumn{2}{c}{Adversarial} \\
              &       & Accuracy & F1 score & Accuracy & F1 score & Accuracy & F1 score \\
        \midrule
        \multirow{10}[2]{*}{LLaVA-1.5} & Woodpecker$\dagger$ & 70.82  & 59.73  & 68.62  & 58.53  & 68.49  & 58.07  \\
              & LURE$\dagger$  & 71.10  & 60.08  & 69.17  & 58.63  & 69.16  & 58.34  \\
              & Greedy & 70.55  & 58.75  & 68.93  & 57.42  & 67.91  & 56.64  \\
              & Beam Search & 71.32  & 60.38  & 69.31  & 58.98  & 69.02  & 58.43  \\
              & OPERA & 71.02  & 59.80  & 69.31  & 58.42  & 68.79  & 58.00  \\
              & VCD   & 71.08  & 60.05  & 68.96  & 58.34  & 68.55  & 58.02  \\
              & DoLa  & 70.73  & 59.36  & 69.14  & 58.08  & 68.32  & 57.44  \\
              & SID   & 71.47  & 61.63  & 69.42  & 59.62  & 69.36  & 58.83  \\
              & HALC  & 70.76  & 60.46  & 69.17  & 59.33  & 69.25  & 58.50  \\
              & Ours  & \textbf{72.03 } & \textbf{62.13 } & \textbf{70.18 } & \textbf{60.93 } & \textbf{70.31 } & \textbf{59.20 } \\
        \midrule
        \multirow{10}[2]{*}{MiniGPT4} & Woodpecker$\dagger$ & 68.05  & 53.84  & 65.49  & 51.70  & 65.06  & 51.27  \\
              & LURE$\dagger$  & 68.12  & 53.91  & 65.96  & 52.37  & 65.17  & 51.38  \\
              & Greedy & 68.02  & 53.71  & 65.31  & 51.68  & 65.41  & 51.92  \\
              & Beam Search & 68.26  & 53.97  & 66.02  & 52.27  & 65.55  & 51.93  \\
              & OPERA & 67.73  & 53.08  & 65.37  & 51.32  & 65.19  & 51.20  \\
              & VCD   & 67.96  & 53.26  & 65.61  & 51.50  & 65.02  & 51.07  \\
              & DoLa  & 68.08  & 53.83  & 65.55  & 51.93  & 65.25  & 51.72  \\
              & SID   & 68.09  & 53.86  & 65.69  & 51.98  & 65.28  & 51.77  \\
              & HALC  & 68.18  & 53.93  & 65.83  & 52.06  & 65.31  & 51.80  \\
              & Ours  & \textbf{68.55 } & \textbf{54.87 } & \textbf{66.23 } & \textbf{52.93 } & \textbf{65.91 } & \textbf{52.70 } \\
        \midrule
        \multirow{10}[2]{*}{ mPLUG-Owl2} & Woodpecker$\dagger$ & 68.61  & 58.10  & 67.28  & 53.07  & 66.58  & 55.42  \\
              & LURE$\dagger$  & 68.78  & 58.28  & 67.35  & 53.15  & 66.89  & 55.65  \\
              & Greedy & 69.67  & 57.40  & 68.02  & 53.43  & 67.14  & 55.43  \\
              & Beam Search & 68.79  & 55.31  & 66.92  & 52.89  & 65.90  & 53.12  \\
              & OPERA & 69.08  & 55.70  & 67.37  & 53.41  & 66.43  & 53.66  \\
              & VCD   & \textbf{70.49 } & \textbf{58.63 } & 68.55  & 54.87  & 67.31  & 56.13  \\
              & DoLa  & 69.61  & 57.21  & 67.90  & 53.38  & 67.08  & 55.24  \\
              & SID   & 69.34  & 55.82  & 67.80  & 53.46  & 67.01  & 56.07  \\
              & HALC  & 69.66  & 56.29  & 67.67  & 53.38  & 66.95  & 55.84  \\
              & Ours  & 70.38  & 58.27  & \textbf{68.70 } & \textbf{55.28 } & \textbf{67.86 } & \textbf{56.77 } \\
        \bottomrule
        \end{tabular}%
      \caption{Comparison of the average Accuracy and F1-score evaluation results under different settings (i.e., \textit{ Random, Popular, Adversarial}) with different baselines and our VASparse on offline POPE benchmark~\cite{Li2023EvaluatingOH, Chen2024HALCOH} of five random runs. Higher F1-score indicate better performance and bold indicates the best results. $\dagger$ denotes the post-hoc method. We set the maximum generated length to 64.}
      \label{tab:opope-appendix-64}%
    \end{table*}%

\section{More evidence of empirical observations}
We present additional evidence on the attention sparsity and attention sinking of LLaVA-1.5 in Figure~\ref{heads_simple}. Our research findings confirm that the self-attention in most layers of the LVLM decoder is sparse. Furthermore, we observe a significant attention "sinking" effect on certain text tokens within the LVLM's attention mechanisms. These results further confirm the characteristics of attention sparsity and attention sinking in LVLMs.

\section{More results on CHAIR benchmark}
We set the maximum generation length to 512 and evaluated our method using the CHAIR benchmark, as shown in Table~\ref{tab:chair-Appendix}. We can observe that when setting the maximum generation length to 512, our method still outperforms the baseline method in most metrics, while achieving competitive decoding speed.
For all results, we set different random seeds and run them five times, and record the average of the results from the five runs.

\section{More results on POPE benchmark}
Following HALC~\cite{Chen2024HALCOH}, we utilize offline POPE (OPOPE) benchmark with both accuracy and F1-score as evaluation metrics to assess VH. 
We conduct experiments under two different maximum text length settings: 64 and 512 tokens.
As shown in Tables~\ref{tab:opope-appendix-512} and~\ref{tab:opope-appendix-64}, we observe several key findings:
(1) VASparse consistently achieves optimal performance across most experimental settings, surpassing both state-of-the-art decoding methods and post-processing approaches under both the 64 and 512-token settings.
(2) The effectiveness of VASparse remains robust across different text length configurations. The performance improvements persist when extending the maximum text length from 64 to 512 tokens, indicating the method's scalability;
(3) VASparse demonstrates consistent VH mitigation capabilities across three distinct LVLM architectures, highlighting its versatility and plug-and-play nature. This architectural agnosticism suggests broad applicability across different model frameworks.

\section{Qualitative Study}
\begin{figure*}[t]
  \centering
  \includegraphics[width=0.97\textwidth]{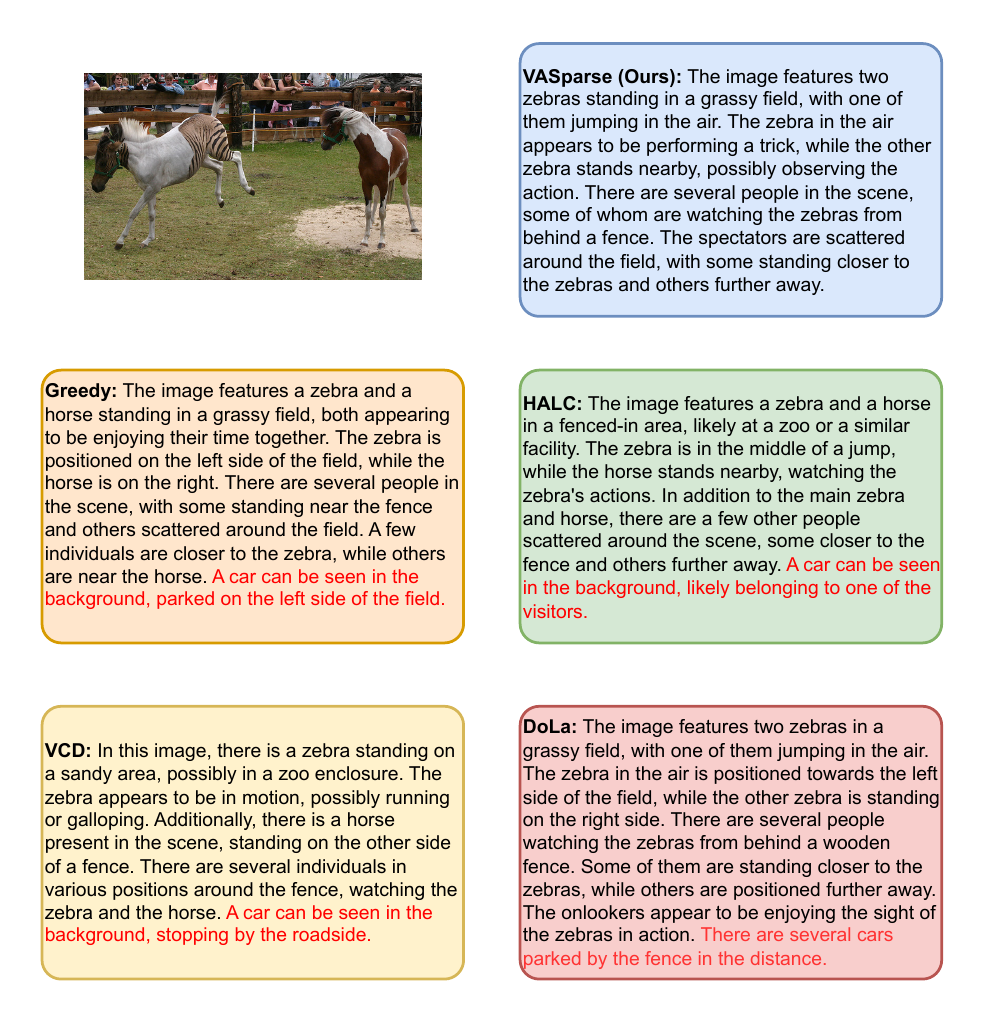}
  \caption{ Qualitative results comparing our VASparse and other methods with LLaVA-1.5 backbone.}
  \label{fig:main-vis-1}
\end{figure*}

\begin{figure*}[t]
  \centering
  \includegraphics[width=0.97\textwidth]{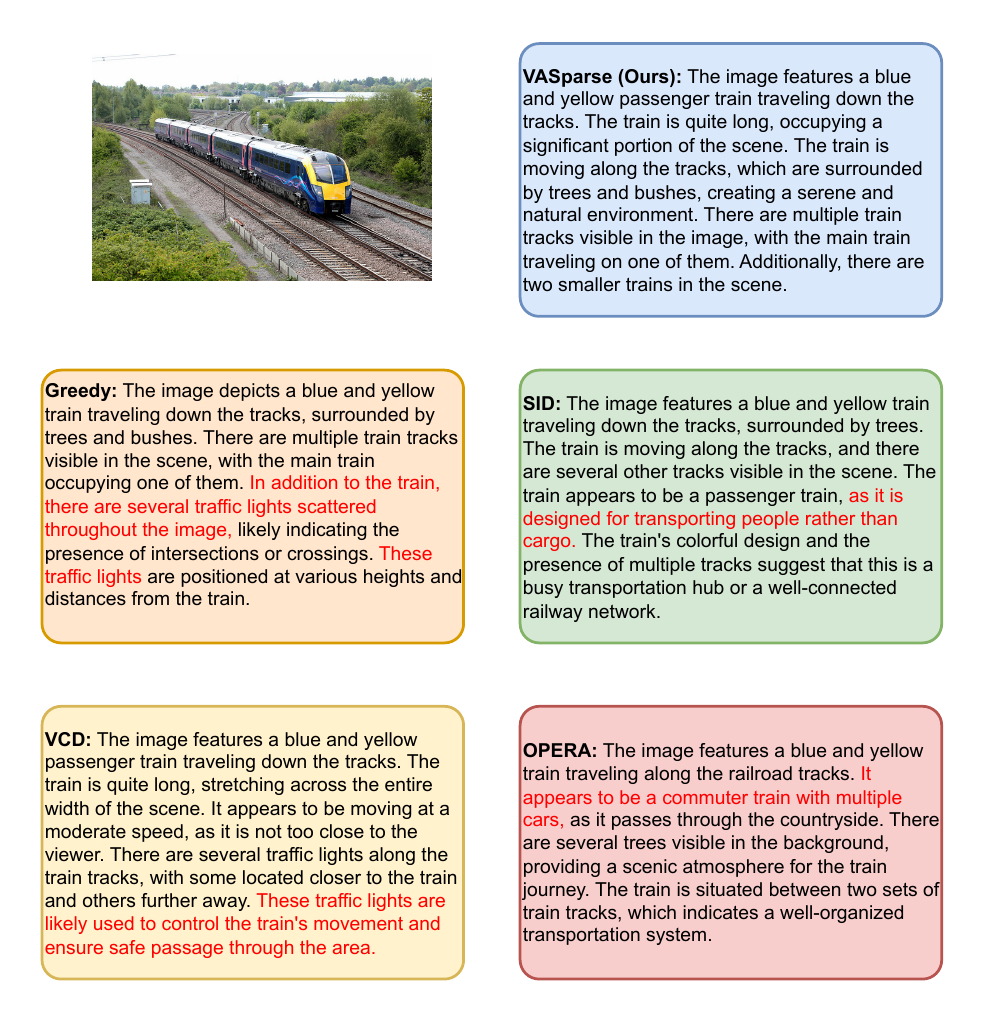}
  \caption{ Qualitative results comparing our VASparse and other methods with LLaVA-1.5 backbone.}
  \label{fig:main-vis-2}
\end{figure*}

\begin{figure*}[t]
  \centering
  \includegraphics[width=0.97\textwidth]{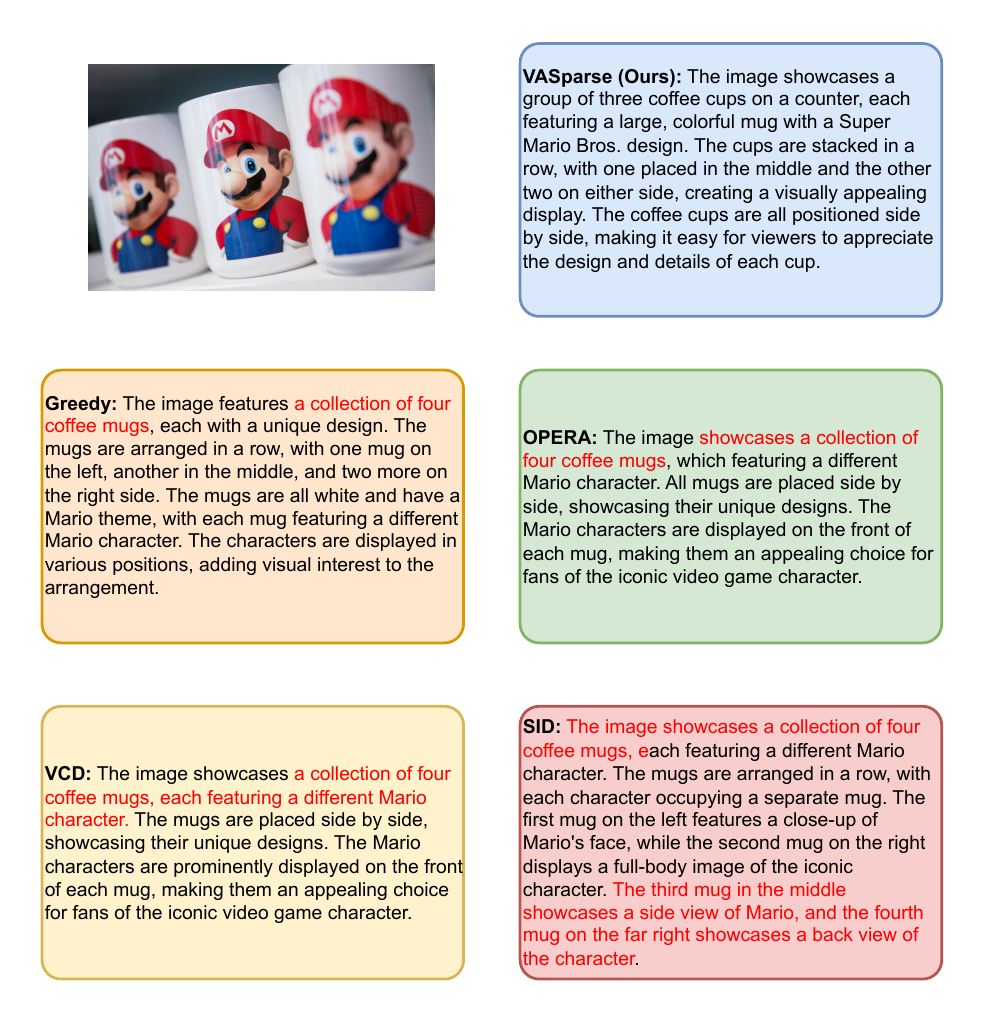}
  \caption{ LLaVA-Bench results comparing our VASparse and other methods with LLaVA-1.5 backbone.}
  \label{fig:main-vis-4}
\end{figure*}

\begin{figure*}[t]
  \centering
  \includegraphics[width=0.97\textwidth]{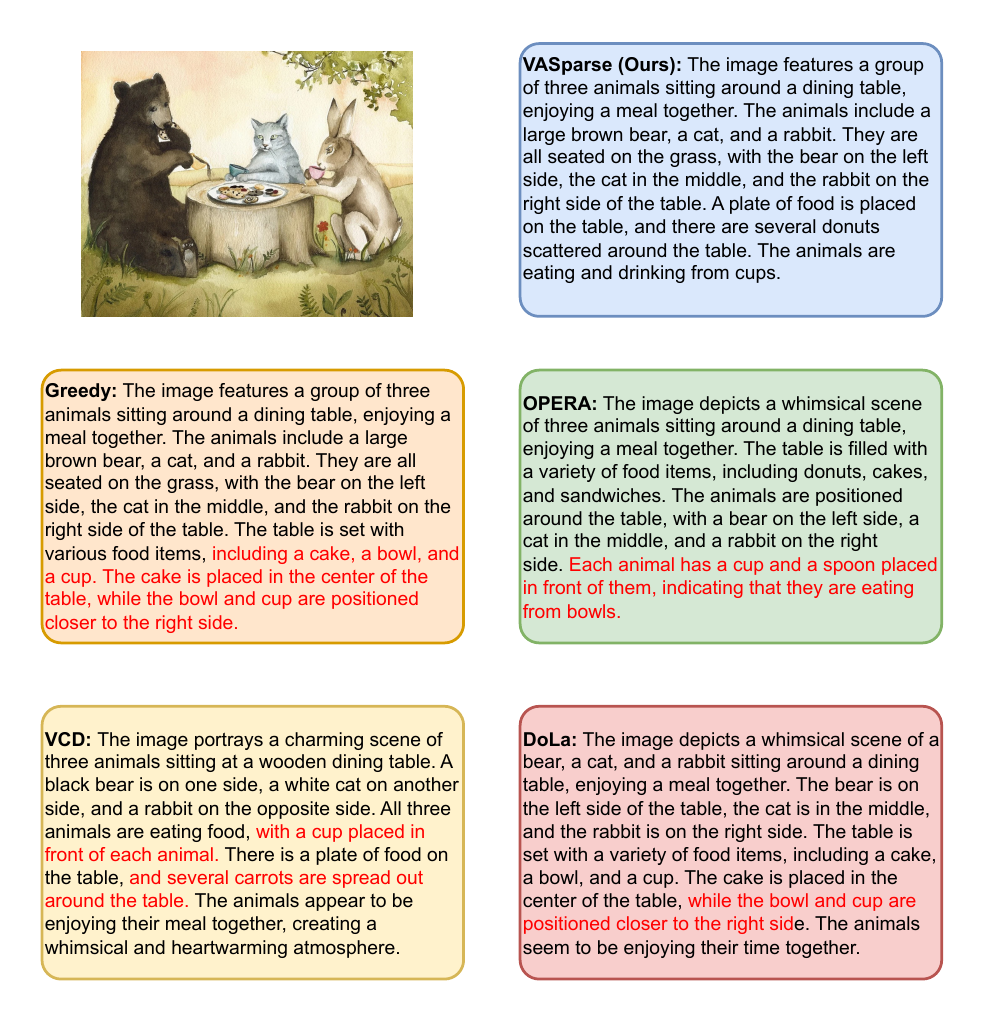}
  \caption{ LLaVA-Bench results comparing our VASparse and other methods with LLaVA-1.5 backbone.}
  \label{fig:main-vis-5}
\end{figure*}

\begin{figure*}[t]
  \centering
  \includegraphics[width=0.97\textwidth]{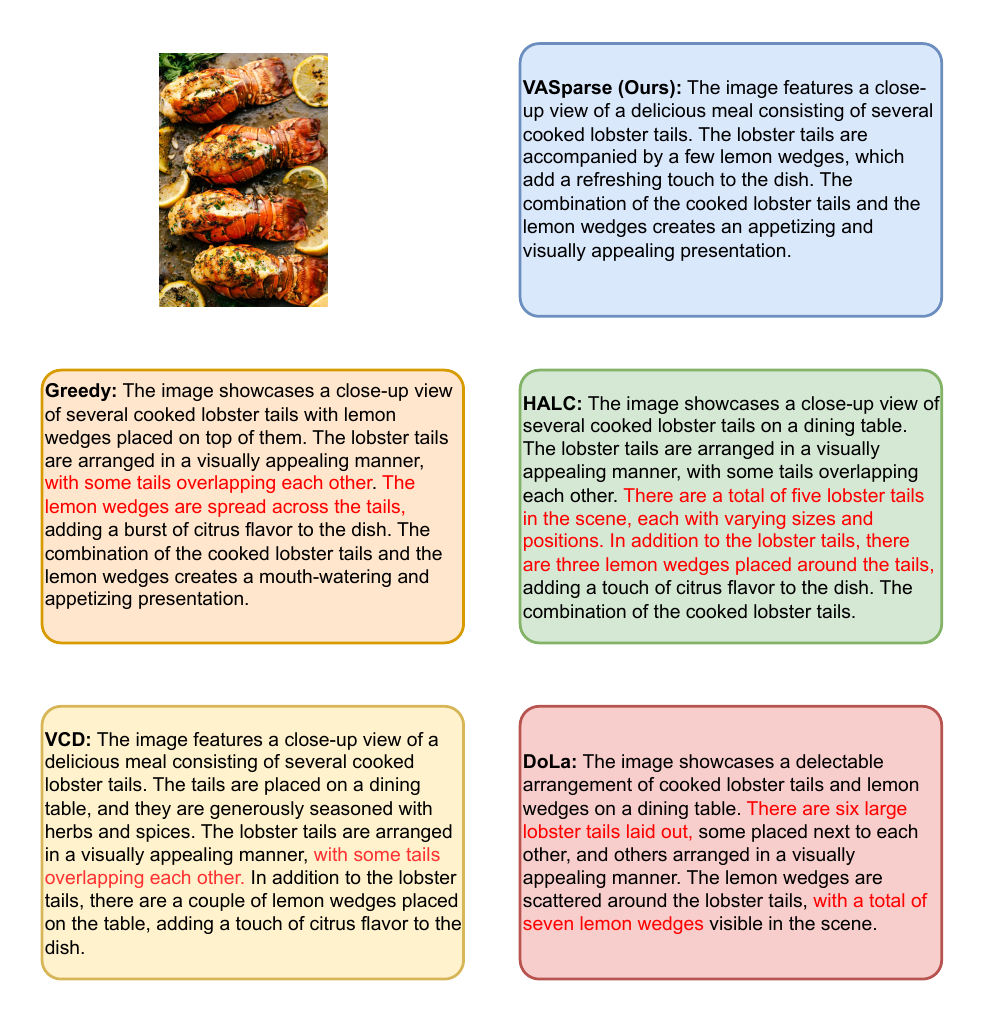}
  \caption{ LLaVA-Bench results comparing our VASparse and other methods with LLaVA-1.5 backbone.}
  \label{fig:main-vis-6}
\end{figure*}

To visually demonstrate the effectiveness of our approach, we present generated captions from our method and baseline approaches in Figures~\ref{fig:main-vis-1} and ~\ref{fig:main-vis-2}on the MSCOCO dataset. 
We consistently used \textit{Please describe this image in detail.} as the input prompt across all experiments. 
The results indicate that captions generated by our VASparse method exhibit notably fewer hallucinated descriptions.
To further evaluate our method's effectiveness in mitigating VH, we conducted experiments on LLaVA-Bench~\cite{Liu2023ImprovedBW}, which consists of 24 distinct images with expert-annotated descriptions and corresponding evaluation questions. 
In alignment with previous studies~\cite{Yin2023WoodpeckerHC, Leng2023MitigatingOH, Chen2024HALCOH}, we employed this benchmark for qualitative assessment of VH reduction. 
The visual results are presented in Figure~\ref{fig:main-vis-4}, ~\ref{fig:main-vis-5} and ~\ref{fig:main-vis-6}, where we consistently used the prompt \textit{Please describe this image in detail.} to generate image captions.

\end{document}